\documentclass[runningheads]{llncs}

\usepackage{eccv}
\usepackage{eccvabbrv}
\usepackage{graphicx}
\usepackage{booktabs}
\usepackage{wrapfig}

\definecolor{MyDarkBlue}{rgb}{0.08,0.02,0.5}
\definecolor{MyLightBlue}{rgb}{0,0.08,0.8}

\newcommand{\supp}[1]{{\color{black}#1}}

\usepackage{colortbl}
\usepackage{multirow}
\usepackage{arydshln}
\usepackage[T1]{fontenc}
\usepackage{listings}
\usepackage{float}
\newfloat{algorithm}{t}{loa}
\floatname{algorithm}{Algorithm}

\definecolor{codeblue}{rgb}{0.25,0.5,0.5}
\definecolor{codefunc}{rgb}{0.85, 0.18, 0.50}

\lstdefinelanguage{PythonV2V}{
  language=Python,
  keywordstyle=\color{blue}\bfseries,
  commentstyle=\color{codeblue},
  stringstyle=\color{orange},
  showstringspaces=false,
  literate=
    {max_cosine_sim} {{\color{codefunc}max\_cosine\_sim}} {15}
    {cosine_sim} {{\color{codefunc}cosine\_sim}} {10}
    {randn_like} {{\color{codefunc}randn\_like}} {10}
    {patchify} {{\color{codefunc}patchify}} {8}
    {slide_kv} {{\color{codefunc}slide\_kv}} {8}
    {denoise} {{\color{codefunc}denoise}} {7}
    {encode(} {{\color{codefunc}encode}(} {7}
    {decode(} {{\color{codefunc}decode}(} {7}
    {chunk} {{\color{codefunc}chunk}} {5}
    {stack} {{\color{codefunc}stack}} {5}
    {topk} {{\color{codefunc}topk}} {4}
    {zeros} {{\color{codefunc}zeros}} {5}
}

\newcommand{\impmark}[1]{{\color{ForestGreen}$\downarrow$#1\%}}
\newcommand{\impmarkup}[1]{{\color{ForestGreen}$\uparrow$#1\%}}

\newcommand{\myparagraph}[1]{\paragraph{#1}}

\usepackage[pagebackref,breaklinks,colorlinks,citecolor=eccvblue]{hyperref}

\title{Consistent Video-to-Video Translation \\ via Explicit Correspondences}

\titlerunning{Consistent Video-to-Video Translation via Explicit Correspondences}
\author{
Gaurav Parmar\inst{1,2} \quad
Zhengqi Li\inst{2} \quad
Richard Zhang\inst{2} \quad
Jun-Yan Zhu\inst{1} \quad  \\
Srinivasa Narasimhan\inst{1} \quad
Eli Shechtman\inst{2} \quad
Yotam Nitzan\inst{2}
}

\authorrunning{Parmar et al.}
\institute{Carnegie Mellon University\inst{1}\; \;
Adobe Research\inst{2}\; \;  }

\begin{document}

\maketitle

\begin{abstract}
Interactive video-to-video applications require real-time generation while maintaining long-range temporal consistency.
However, recent methods achieve speed by restricting attention to a fixed-length causal window. This limits long-range consistency whenever relevant history falls outside the context window.
To address this limitation, we introduce \texttt{vid2vid-long}, a correspondence-based approach grounded in a simple observation.
Video-to-video translation provides an \emph{explicit} cross-frame reference: the regions that match in the input video should match in the output video too.
Concretely, we compute patch-level correspondences in the input, guiding retrieval of previously generated outputs in the corresponding regions, 
and condition the model on these patches through correspondence cross-attention.
Our experiments demonstrate that \texttt{vid2vid-long} consistently improves long-range temporal consistency across recent video models while preserving generation quality and real-time performance.

\end{abstract}
\vspace{-5mm}

\begin{figure}[t]
    \centering 
    \includegraphics[width=\linewidth]{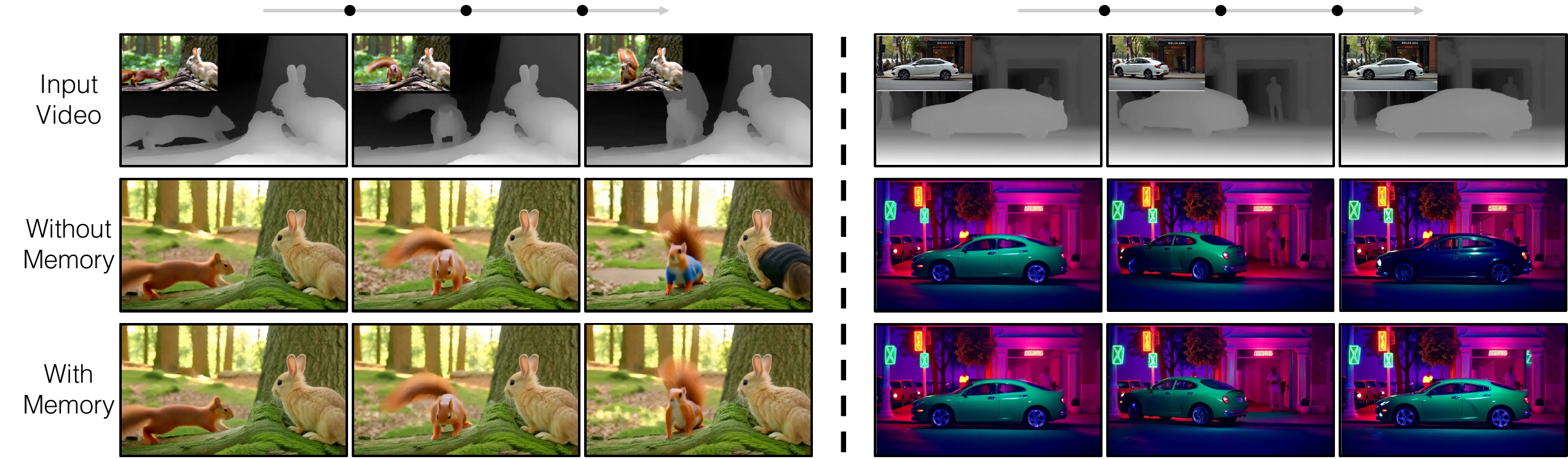}
    \caption{
    \textbf{Long-range consistency without long context windows.} 
    Standard auto-regressive video translation forgets content once it falls outside its context window (middle row). Our proposed method \texttt{vid2vid-long} (bottom row) explicitly retrieves matching past outputs when the inputs recur, maintaining consistency over long videos. \supp{Please see the supplement for a large grid of qualitative result videos.}
    }
    \label{fig:teaser}
\end{figure}

\section{Introduction}
\label{sec:intro}
Real-time video-to-video translation unlocks a large class of interactive applications, including streaming, gaming, and augmented and virtual reality.
Recent advances in large scale text-to-video diffusion models~\cite{wan2025wan, opensora, opensora2, blattmann2023stable, kong2024hunyuanvideo,gao2025seedance10exploringboundaries}, and subsequently 
distilling them into few-step, autoregressive models~\cite{yin2025causvid, huang2025self, yang2025longlive, liu2025rolling, lin2025diffusionadversarialposttrainingonestep}, have enabled high-quality generation at interactive speeds.

Autoregressive video generation relies on a history of past frames to ensure temporal continuity. However, determining an efficient mechanism for maintaining this history remains a fundamental challenge.
Na\"ively retaining all past outputs causes memory to scale linearly and runtime quadratically with sequence length, both of which are prohibitive for real-time applications.
To maintain real-time latency, many methods resort to a fixed-length sliding window.
While this ensures a fixed memory and runtime per frame, it fundamentally sacrifices long-range consistency: once an object leaves the window, often after only a few seconds, the model ``forgets'' it forever.

Several lines of work have attempted to bridge this gap. First, attention sinks~\cite{xiao2023streamingllm, yang2025longlive,liu2025rolling,shin2026motionstream} heuristically retain the initial frames as a global anchor.

However, this approach inherently cannot address novel content that first appears in a later frame, even in cases like simple disocclusions. Another approach involves models with learned hidden states, such as Mamba~\cite{gu2024mambalineartimesequencemodeling, dao2024transformersssmsgeneralizedmodels}, which compress history into fixed-size bottlenecks. While efficient, these models often forgo the expressive power of transformers and remain largely limited to scenes with lower temporal complexity~\cite{po2025longcontextstatespacevideoworld}. 

Finally, some methods use pose conditioning~\cite{yu2025cam} or explicit world representation (e.g., 3D point clouds) to maintain a consistent world~\cite{xiao2025worldmemlongtermconsistentworld, wang2025evoworldevolvingpanoramicworld}. These methods struggle to model dynamic scenes and require additional privileged inputs that are hard to obtain. 

In this paper, we address the challenge of long-range consistency in video-to-video translation.
Our insight is that long-range consistency cues are provided in the input video itself. 
That is, if two regions match in the input video, their generated outputs should match as well.
The input space can therefore serve as an index for retrieving previously generated content, allowing consistency to be enforced, without requiring access to the full output history.

As such, we introduce \texttt{vid2vid-long}, a correspondence-based dynamic memory mechanism that enables long-term consistency while preserving real-time performance.
Concretely, we compute patch-level correspondences in the input video and use them to index a compressed global representation of prior outputs.
This representation grows only when novel content appears, rather than with video length, resulting in sub-linear memory growth in practice.
The approach is complementary to existing autoregressive distillation backbones.

We evaluate our approach in combination with multiple autoregressive distillation frameworks, including Self-Forcing, LongLive, and Rolling Forcing. 
In all settings,
incorporating
our technique substantially improves long-range consistency, while maintaining real-time inference speed and high visual quality.
Our main contributions are:
\begin{itemize}
    \item We identify a structural property of video-to-video translation: correspondences in input space induce correspondences in output space, enabling input-driven retrieval for long-range consistency.
    \item We propose a correspondence-based mechanism that maintains a compressed global representation of prior outputs whose size scales with visual novelty rather than sequence length.
    \item We demonstrate consistent improvements in long-range temporal coherence across multiple real-time autoregressive backbones.
\end{itemize}
Our project website is available at \url{https://www.cs.cmu.edu/~vid2vid-long/}.

\begin{figure}[t!]
    \centering 
    \includegraphics[width=\linewidth]{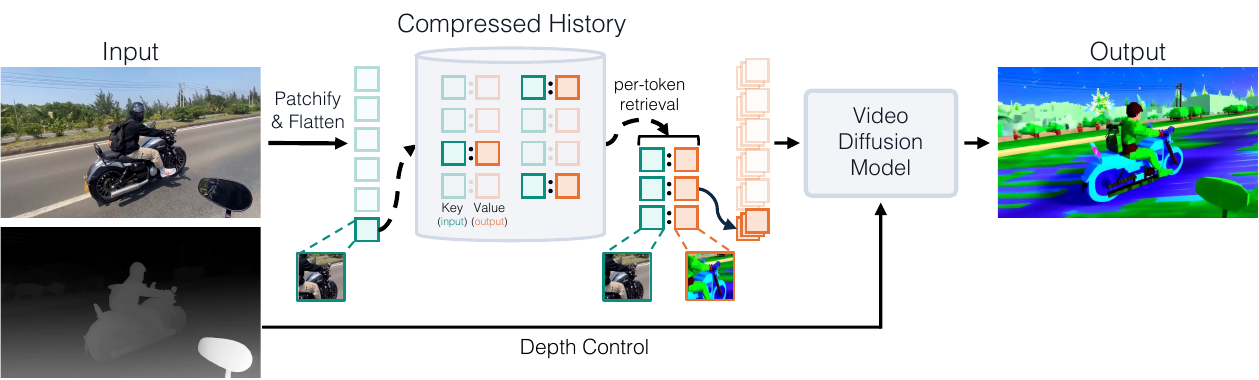}
    \caption{
    \textbf{Correspondence-based video translation.}
    The memory is populated with input-output correspondence pairs from previous generations. To support the consistent translation of the next block of frames, we first retrieve from memory the output patches whose corresponding input is most similar to the current.
    These output patches, together with the normal input signal, is the input to our video diffusion model.
    Based on the new induced input-output pairs, we write to memory only those that are sufficiently different from existing. 
    }
    \label{fig:arch}
\end{figure}

\section{Related Works}
\label{sec:related_works}

\myparagraph{Controllable Generation.}
The rapid advancements in text-to-image~\cite{rombach2022high, flux2024, wu2025qwenimagetechnicalreport} and text-to-video~\cite{blattmann2023stable, wan2025wan, HaCohen2024LTXVideo} have been accompanied by controllable adapters that allow a more fine-grained control over the generation. 
In the image domain, works like \textit{ControlNet}~\cite{zhang2023adding}, \textit{Sketch-Guided-Diffusion}~\cite{voynov2023sketch}, \textit{T2I adapter}~\cite{mou2023t2i}, \textit{img2img-turbo}~\cite{parmar2024one} show the ability to add structure control like edge maps, depth maps, and hand drawn sketches. 
Models such as \textit{IP-adapter}~\cite{ye2023ip-adapter}, \textit{BLIP-diffusion}~\cite{li2023blip}, \textit{SynCD}~\cite{kumari2025syncd}, \textit{VisualComposer}~\cite{parmar2025object} show ability to condition the generation of subject identity. 
These methods have also been extended to videos in recent methods like \textit{ControlVideo}~\cite{zhang2023controlvideo}, \textit{Control-A-Video}~\cite{chen2023control}, \textit{VACE}~\cite{vace}, and COSMOS-T~\cite{nvidia2025worldsimulationvideofoundation}. 
However, existing video control methods are primarily designed for offline generation, often relying on non-causal processing that makes them unsuitable for interactive applications.
We aim to address this gap in this paper. 

\vspace{-0.1in}
\myparagraph{Diffusion Distillation.}
Several prior works have explored speeding up diffusion models by reducing the number of denoising steps. Initial image domain approaches like \textit{consistency models}~\cite{song2023consistency}, \textit{progressive distillation}~\cite{salimans2022progressive}, \textit{adversarial distillation}~\cite{sauer2024adversarial}, and distribution matching distillation~\cite{yin2024one, yin2024improved}, \textit{moment matching distillation}~\cite{salimans2024multistep} enabled high-quality inference in 1--4 steps. 
Recent methods like \textit{CausVid}~\cite{yin2025causvid} extend distillation to video generation, while \textit{Diffusion Forcing}~\cite{chen2024diffusion} explores autoregressive video generation. However, these approaches can still suffer from error accumulation during autoregressive inference.
To address this, \textit{Self-Forcing}~\cite{huang2025self} trains the student model on its own generated rollout, rather than ground truth video.
Our work is complementary to these methods, focusing on maintaining long-range consistency during autoregressive inference.

\vspace{-0.1in}
\myparagraph{Long Context Generation.}
Maintaining long-term consistency in autoregressive generation is computationally challenging due to the quadratic complexity of transformer attention.
This is particularly severe for videos: five seconds of 480p video can exceed 32,000 tokens in modern architectures like WAN2.1~\cite{wan2025wan}.
Existing methods such as \textit{Streaming-T2V}~\cite{henschel2025streamingt2v} or \textit{LongLive}~\cite{yang2025longlive} typically use sliding windows or hierarchical latent memory.
However, these often suffer from recency bias, where the model forgets the appearance of objects or environments during loop closures or long-duration occlusions.
While the language modeling literature has explored KV cache pruning and compression~\cite{ge2023model, liu2025chunkkv}, these methods often ignore the spatiotemporal continuity of video.

Rather than deciding which history to retain solely based on temporal position, learned compression, or token importance, we use input-space correspondences to retrieve the specific past outputs that are relevant to the current frame. This makes the memory mechanism task-aware and particularly well-suited to video-to-video translation, where repeated input regions should induce repeated output appearances.

\section{Method}
\label{sec:method}
Our objective is to design an autoregressive video-to-video diffusion model equipped with a correspondence-based memory mechanism. Our architecture enables long-range temporal consistency while maintaining the low-latency performance required for real-time deployment. We begin by reviewing the autoregressive formulation in Section~\ref{sec:preliminaries}. In Section~\ref{sec:history}, we introduce our explicit correspondence-based memory component. We then detail our strategy for conditioning on this historical context in Section~\ref{sec:correspondence}. Finally, we describe the specific training objectives and real-time inference algorithm in Sections~\ref{sec:training} and~\ref{sec:inference}, respectively.

\subsection{Preliminaries}
\label{sec:preliminaries}

\myparagraph{Bidirectional Video Generation.}
\label{sec:prelim_bid}
Our method builds upon pre-trained flow-matching models designed for video-to-video translation. Specifically, we adopt VACE~\cite{vace}, which is a latent flow-matching model, operating within the latent space of a pre-trained Variational Autoencoder (VAE)~\cite{rombach2022high}.
Given a block of input frames $\mathbf{I}^{(b)} \in \mathbb{R}^{T \times H \times W \times 3}$, the VAE encoder $\mathcal{E}$ maps the pixel data to a latent representation $\mathbf{z}^{(b)} \in \mathbb{R}^{t \times h \times w \times c}$.
The latent dimensions are determined by the spatial and temporal compression factors, which for VACE are given as follows: $h = H/8$, $w = W/8$, and $t = 1 + (T-1)/4$.
Following common practice in vision transformers~\cite{peebles2023scalable}, the latent block is further patchified into a sequence of tokens.
Each latent frame is partitioned spatially into non-overlapping patches of size $1 \times p \times p$, where the temporal dimension of a patch remains singular.
This process yields a total of $N = t \times (h/p) \times (w/p)$ tokens per block. 
Specifically for VACE, $p=2$.

For the generative model itself, VACE utilizes a bidirectional spatiotemporal transformer to generate an entire video at once.
This design implies that every generated token depends on the full context of the video, including both past and future tokens.
This design is expensive in computation and memory, limiting their operation to short, fixed-length sequences, typically up to five seconds.

\myparagraph{Autoregressive Video Generation.}\label{sec:prelim_ar}
To enable long-form generation, recent methods~\cite{yin2025causvid, huang2025self, yang2025longlive, liu2025rolling} replace bidirectional attention with causal attention. 
These models generate videos autoregressively in temporal blocks, where each block $\mathbf{z}^{(b)}$ is generated by conditioning on the previously generated blocks $\mathbf{z}^{(<b)}$.
Access to previously generated blocks is typically implemented via a causal KV cache.
While theoretically sound, maintaining the entire history $\mathbf{z}^{(<b)}$ causes the runtime to grow quadratically with the video length, and the memory usage linearly.
To ensure a constant per-frame runtime necessary for streaming applications, the  causal attention is typically restricted to a smaller set of past frames, implemented with a sliding window of size $W$.
Under this constraint, the generation of block $b$ conditions only on the most recent context $\mathbf{z}^{(b-W:b-1)}$.
Although this approach bounds the computational cost, it fundamentally limits temporal consistency to the window size. 
Any information that falls outside this window becomes inaccessible to the model, leading to identity drift and ``forgetting'' over long horizons.
Our method addresses this limitation by providing an explicit correspondence-based memory that allows any autoregressive model to access the history arbitrarily far beyond its fixed window.

\subsection{Compressed Patch History}
\label{sec:history}

In video-to-video translation, the availability of the input stream allows us to move beyond temporal proximity as the sole proxy for relevance.
While a sliding window approach utilizes only recent frames for context, we can use explicit correspondences in the input space to identify relevant history from any point in the past.
This is a reliable strategy because regions that \emph{match} in the input video should \emph{match} in the generated output (see Figure~\ref{fig:arch}).

By explicitly hard-coding these correspondences, we offload the burden of discovery from the transformer's expensive attention layers.
Rather than forcing the model to rediscover these relationships, we provide them directly via an external index. 
We represent this memory as an evolving set of input-output pairs.
Each entry consists of a descriptor for an input patch, which acts as a retrieval key, and the corresponding generated output patch, which serves as the value for future conditioning.
To ensure the system remains scalable for long sequences, we only update the memory when the input video presents novel information. 
If an incoming patch is already sufficiently represented in the history, we do not save it.
This ensures that the memory footprint scales with the visual novelty of the scene, rather than the raw sequence length.
\supp{(See history growth analysis in the appendix.)}

To implement this mechanism, we define three concrete operations that govern the memory lifecycle. We first specify the patch-level representation for the keys and values, followed by the retrieval protocol for reading from history, and finally the novelty-based policy for writing to the global store.

\vspace{-2mm}
\subsubsection{Patch Descriptor.}
Our memory mechanism leverages the spatial discretization already present in the diffusion transformer (see Section~\ref{sec:prelim_bid}). 
In this framework, every input video spatiotemporal patch is already mapped to a small latent patch of shape $(1,2,2)$ that corresponds to a single token.
From here, we use term ``latent patch'' for these token-corresponding patches.
We assign each latent patch a global index $g$ that uniquely identifies its spatiotemporal coordinates. 
For every input latent patch at index $g$, denoted as $\mathbf{z}_g$, we define a descriptor $\mathbf{k}_g$ to serve as a retrieval key.
We evaluated several feature extractors $\phi$ to produce these keys, including external vision backbones applied in pixel space. 
However, our experiments show that the most effective representation is the latent patch itself (further analysis is included in the appendix).
Consequently, we define the feature extraction function as the identity mapping, where $\mathbf{k}_g = \mathbf{z}_g$. 
This approach also avoids the computational cost of decoding latents back to pixels and the inference run of additional feature extraction backbones. 
Under this formulation, the memory stores pairs $(\mathbf{k}_g, \mathbf{v}_g)$, where the key $\mathbf{k}_g$ is the input latent patch and the value $\mathbf{v}_g$ is the corresponding to the generated output latent patch.

\subsubsection{Memory Management.}
To maintain temporal coherence throughout the generation process, we define the memory as a dynamic set of key-value pairs that evolves alongside the autoregressive blocks.
Let $\mathcal{H}_b$ represent the memory state after the generation of block $b$, initialized as an empty set $\mathcal{H}_0 = \emptyset$.
We similarly define $\mathcal{G}_b$ as the set of all global indices $g$ associated with the latent tokens generated within block $b$.
Under this formulation, the generation of each subsequent block $b+1$ utilizes the existing history $\mathcal{H}_b$ for retrieval and concludes by updating the state to $\mathcal{H}_{b+1}$ using the newly generated tokens in $\mathcal{G}_{b+1}$.
\myparagraph{Memory Write Policy.}
The memory state is updated after the generation of each block by evaluating the novelty of every new latent patch relative to the existing stored patches. 
For each generated latent patch at index $g \in \mathcal{G}_{b+1}$, we pair the input descriptor $\mathbf{k}_g$ with the corresponding output  latent patch $\mathbf{v}_g$.
The similarity between a query latent patch $\mathbf{k}_g$ and the current memory $\mathcal{H}_b$ is defined as the maximum cosine similarity:
\begin{equation}
\text{sim}(\mathbf{k}_g, \mathcal{H}_b) = \max_{(\mathbf{k}_{g'}, \mathbf{v}_{g'}) \in \mathcal{H}_b} \text{sim}(\mathbf{k}_g, \mathbf{k}_{g'}) = \max_{(\mathbf{k}_{g'}, \mathbf{v}_{g'}) \in \mathcal{H}_b}\frac{\mathbf{k}_g \cdot \mathbf{k}_{g'}}{\|\mathbf{k}_g\| \|\mathbf{k}_{g'}\|}.
\end{equation}
For an empty memory, we define \(\operatorname{sim}(k_g,\mathcal{H}_0)=-\infty\), ensuring that all patches from the first generated block are added to memory.
The latent patch and its corresponding generated output $\mathbf{v}_g$ are committed to the memory only if this similarity falls below a novelty threshold $\tau_h$. The updated memory $\mathcal{H}_{b+1}$ is thus defined as:
\begin{equation}
\label{eq:memory_write}
\mathcal{H}_{b+1} = \mathcal{H}_b \cup \{ (\mathbf{k}_g, \mathbf{v}_g) : \text{sim}(\mathbf{k}_g, \mathcal{H}_b) < \tau_h \}_{g \in \mathcal{G}_{b+1}} .
\end{equation}

This policy ensures that memory footprint and inference cost scales with discovery of unique visual content rather than total sequence length.

\myparagraph{Memory Read Mechanism.}
Before generating the output latents for block $b+1$, we query $\mathcal{H}_b$ using the descriptors of the current input latent. For each patch index $g$, we identify the set of $k$ nearest neighbors $\mathcal{N}_k(g)$ whose similarity exceeds a retrieval threshold $\tau_m$:
\begin{equation}
\label{eq:memory_read}
\vspace{-1mm}
\mathcal{N}_{k}(g) = \operatorname{Top}_k 
( \{ (\mathbf{k}_{g'}, \mathbf{v}_{g'})
\in \mathcal{H}_b : \text{sim}(\mathbf{k}_g, \mathbf{k}_{g'}) \ge \tau_m 
\}).
\vspace{-1mm}
\end{equation}
For every query latent patch that satisfies this condition, we retrieve the stored output values ${\mathbf{v}_{g'}}$. If no entries in $\mathcal{H}_b$ meet the threshold $\tau_m$, the neighborhood $\mathcal{N}_k(g)$ is empty. 
We track these occurrences with a binary validity mask $\mathbf{m}_g$, where $\mathbf{m}_g = 1$ if $\mathcal{N}_k(g) \neq \emptyset$ and $0$ otherwise. This mask and the retrieved values are then used to condition the diffusion transformer as described in Section~\ref{sec:correspondence}.

\subsection{Conditioning via Correspondence Cross-Attention}
\label{sec:correspondence}

We now describe how these retrieved values are used to condition the video translation model.
The retrieved output patches $\{\mathbf{v}_{g'}\}$ are tokenized using the same patch embedding as the input latents, producing history tokens $\mathbf{h}_{g'}$ in a shared representation space.

To incorporate these history tokens into the video-model architecture, we add correspondence cross-attention layers to every transformer block between self-attention and text cross-attention layers. 
Each token attends only to its own $k$ retrieved neighbors.
Concretely, for each token index $g$, the current hidden state $\mathbf{x}_g$ provides the query.
Let $H_g = \{ \mathbf{h}_{g'} \}_{(\mathbf{k}_{g'}, \mathbf{v}_{g'}) \in \mathcal{N}_k(g)}$ 
denote the set of retrieved history tokens for query $g$. We define
\begin{equation}
\label{eq:corr_xa}
\text{CorrXA}(g) = \mathbf{m}_g \cdot W_o \,
\text{Attn}(W_q \mathbf{x}_g,\; W_k H_g,\; W_v H_g).
\end{equation}

where $\mathbf{m}_g$ is the validity mask that zeros out tokens without valid correspondences.
The output is added as a residual: $\mathbf{x}_g \leftarrow \mathbf{x}_g + \text{CorrXA}(g)$.
The output projection $W_o$ is zero-initialized~\cite{zhang2023adding}, so the layer acts as an identity at the start of training, preserving the behavior of the pretrained model.
Since each token attends only to its $k$ neighbors, the computational overhead of these layers is minimal.
We've described the cross-attention operation for a single token.
To implement this efficiently, we reshape the query and memory tensors to collapse the sequence length into the batch dimension, effectively treating each token's cross-attention as an independent $1 \times k$ operation.

\subsection{Training}
\label{sec:training}
We incorporate our correspondence-based memory into several different autoregressive video model. Unless stated otherwise, we use Rolling Forcing~\cite{liu2025rolling} distillation with 5 steps as the default configuration.
The student model is initialized from pretrained VACE 1.3B and all model parameters are finetuned jointly, including the newly introduced correspondence cross-attention layers.
We adopt the standard Rolling Forcing objective without modification or extra loss terms.
The correspondence memory is active during training, and memory read and write operations are performed as described in Section~\ref{sec:history}.
The student conditions on retrieved correspondences from its own previously generated blocks, and newly generated patches are written to memory according to the novelty threshold.
The teacher model is a bidirectional VACE 14B model and it does not use correspondence memory.

\subsection{Inference}
\label{sec:inference}
At inference time, the model generates the video autoregressively, processing fixed-size blocks sequentially.
At the start of each video, we set $\mathcal{H}_0 = \emptyset$.
For each block $b$, we compute the input latent tokens $\mathbf{z}_g,\ g\in\mathcal{G}_b$.
Using $\mathbf{z}_g$, we query $\mathcal{H}_{b-1}$ to compute neighbors $\mathcal{N}_k(g)$ for each $g \in \mathcal{G}_b$.
The retrieved neighbors are used to condition the translation network through correspondence cross-attention layers during each denoising step of block $b$.
Retrieval is performed once per block and the resulting neighbors are reused across all denoising steps.
After block $b$ is fully denoised, we obtain the output latents $\mathbf{v}_g,\ g \in \mathcal{G}_b$, and update the memory according to Eq.~\ref{eq:memory_write}.
The updated memory $\mathcal{H}_{b}$ is used to generate the next block $b+1$.
A complete inference pseudocode is provided in the appendix.

\begin{table}[t!]
\centering
\caption{\textbf{Quantitative evaluation on depth-to-video translation.} We report long-context reconstruction metrics (LPIPS, DreamSim, CLIP-Sim), quality scores (V-Bench), and inference efficiency. Offline methods are shown in gray on the top and distillation based autoregressive methods are shown at the bottom. The best method in each column is marked in bold and the second best method is marked in underline.}
\resizebox{\textwidth}{!}{%
\begin{tabular}{lllcccccc}
\toprule
\multirow{2}{*}{\textbf{Family}} & \multirow{2}{*}{\textbf{Method}} & \multirow{2}{*}{\shortstack{\textbf{Base} \\ \textbf{Model}}} & \multicolumn{3}{c}{\textbf{Reconstruction}} & \multirow{2}{*}{\shortstack{\textbf{Quality} \\ \textbf{Score}$_\uparrow$}} & \multicolumn{2}{c}{\textbf{Efficiency}} \\
\cmidrule(lr){4-6} \cmidrule(lr){8-9}
& & & \textbf{LPIPS}$_\downarrow$ & \textbf{DreamSim}$_\downarrow$ & \textbf{CLIP-Sim}$_\uparrow$ & & \textbf{Latency(s)}$_\downarrow$ & \textbf{FPS}$_\uparrow$ \\
\midrule

\multirow{4}{*}{\shortstack[l]{Offline}} &{\color{gray} VACE 1.3B} & {\color{gray} N/A} &
{\color{gray} 0.463} & {\color{gray} 0.195} & {\color{gray} 0.901} & {\color{gray} 0.835} & {\color{gray} 79} & {\color{gray} 1.03} \\
& {\color{gray} Control-A-Video} & {\color{gray} SD 1.5} &
{\color{gray} 0.288} & {\color{gray} 0.169} & {\color{gray} 0.907} & {\color{gray} 0.722} & {\color{gray} 11} & {\color{gray} 0.89} \\
& {\color{gray} VideoComposer} & {\color{gray} N/A} &
{\color{gray} 0.664} & {\color{gray} 0.333} & {\color{gray} 0.847} & {\color{gray} 0.724} & {\color{gray} 11} & {\color{gray} 1.36} \\
& {\color{gray} ControlVideo} & {\color{gray} SD 1.5} &
{\color{gray} 0.174} & {\color{gray} 0.065} & {\color{gray} 0.954} & {\color{gray} 0.778} & {\color{gray} 550} & {\color{gray} 0.59} \\
\hdashline

\multirow{5}{*}{\shortstack[l]{Auto- \\ regressive}} & Self-Forcing (4 steps) & VACE 1.3B &
0.548 & 0.203 & 0.881 & 0.789 & 0.67 & 13.74 \\
& LongLive (4 steps) & VACE 1.3B &
0.511 & 0.218 & 0.874 & \underline{0.817} & 0.68 & \underline{13.83} \\
& Rolling Forcing (5 steps) & VACE 1.3B &
0.427 & 0.172 & 0.905 & 0.799 & 2.58 & 10.99 \\

\cdashline{2-9}

& Ours (2 step) & VACE 1.3B &
\underline{0.302} & \underline{0.079} & \underline{0.941} & 0.814 & 1.09 & \textbf{13.88} \\
& Ours (5 step) & VACE 1.3B &
\textbf{0.240} & \textbf{0.066} & \textbf{0.955} & \textbf{0.828} & 3.72 & 8.10 \\
\bottomrule
\end{tabular}
}
\label{tab:memory_rec}
\end{table}

\begin{table*}[ht!]
\centering
\caption{\textbf{Quantitative evaluation on depth-to-video translation (DL3DV).} We report long-context reconstruction metrics (LPIPS, DreamSim, CLIP-Sim), quality scores (V-Bench), and inference efficiency. Offline methods (gray) are shown on the top and distillation based autoregressive methods are in the bottom.The best method in each column is marked in bold and the second best method is marked in underline.}
\resizebox{\textwidth}{!}{%
\begin{tabular}{lllcccccc}
\toprule
\multirow{2}{*}{\textbf{Family}} & \multirow{2}{*}{\textbf{Method}} & \multirow{2}{*}{\shortstack{\textbf{Base} \\ \textbf{Model}}} & \multicolumn{3}{c}{\textbf{Reconstruction}} & \multirow{2}{*}{\shortstack{\textbf{Quality} \\ \textbf{Score}$_\uparrow$}} & \multicolumn{2}{c}{\textbf{Efficiency}} \\
\cmidrule(lr){4-6} \cmidrule(lr){8-9}
& & & \textbf{LPIPS}$_\downarrow$ & \textbf{DreamSim}$_\downarrow$ & \textbf{CLIP-Sim}$_\uparrow$ & & \textbf{Latency(s)}$_\downarrow$ & \textbf{FPS}$_\uparrow$ \\
\midrule

\multirow{4}{*}{\shortstack[l]{Offline}} &
{\color{gray} VACE 1.3B} & {\color{gray} N/A} &
{\color{gray} 0.592} & {\color{gray} 0.240} & {\color{gray} 0.881} & {\color{gray} 0.819} & {\color{gray} 79} & {\color{gray} 1.01} \\
& {\color{gray} Control-A-Video} & {\color{gray} SD 1.5} &
{\color{gray} 0.437} & {\color{gray} 0.188} & {\color{gray} 0.895} & {\color{gray} 0.727} & {\color{gray} 14} & {\color{gray} 0.89} \\
& {\color{gray} VideoComposer} & {\color{gray} N/A} &
{\color{gray} 0.672} & {\color{gray} 0.303} & {\color{gray} 0.854} & {\color{gray} 0.718} & {\color{gray} 12} & {\color{gray} 1.39} \\
& {\color{gray} ControlVideo} & {\color{gray} SD 1.5} &
{\color{gray} 0.205} & {\color{gray} 0.057} & {\color{gray} 0.957} & {\color{gray} 0.794} & {\color{gray} 2518} & {\color{gray} 0.38} \\
\hdashline

\multirow{5}{*}{\shortstack[l]{Auto- \\ regressive}} & Self-Forcing & VACE 1.3B &
0.616 & 0.238 & 0.870 & 0.790 & 0.67 & 13.74 \\
& LongLive & VACE 1.3B &
0.534 & 0.170 & 0.896 & \underline{0.806} & 0.68 & \underline{13.83} \\
& Rolling Forcing & VACE 1.3B &
0.445 & 0.163 & 0.907 & 0.806 & 2.58 & 10.99 \\

\cdashline{2-9}

& Ours (2 step) & VACE 1.3B &
\underline{0.379} & \underline{0.083} & \underline{0.932} & 0.799 & 1.09 & \textbf{13.88} \\
& Ours (5 step) & VACE 1.3B &
\textbf{0.305} & \textbf{0.058} & \textbf{0.952} & \textbf{0.821} & 3.72 & 8.10 \\
\bottomrule
\end{tabular}
}
\label{tab:memory_rec_dl3dv}
\end{table*}

\begin{table}[t]
\centering
\caption{\textbf{Sensitivity to the distillation method.} Our correspondence mechanism consistently improves long-context reconstruction across all three autoregressive distillation methods, with minimal impact on quality. The base model used is VACE 1.3B.}
\resizebox{\textwidth}{!}{%
\begin{tabular}{lcccccccc}
\toprule
& \multicolumn{4}{c}{\textbf{VACE-Bench}} & \multicolumn{4}{c}{\textbf{DL3DV}} \\
\cmidrule(lr){2-5} \cmidrule(lr){6-9}
& \textbf{LPIPS}$_\downarrow$ & \textbf{DreamSim}$_\downarrow$ & \textbf{CLIP-Sim}$_\uparrow$ & \textbf{Quality Score}$_\uparrow$
& \textbf{LPIPS}$_\downarrow$ & \textbf{DreamSim}$_\downarrow$ & \textbf{CLIP-Sim}$_\uparrow$ & \textbf{Quality Score}$_\uparrow$ \\
\midrule

Self-Forcing &
0.548 & 0.203 & 0.881 & 0.789 &
0.616 & 0.238 & 0.870 & 0.790\\
\quad + Correspondence &
0.343\rlap{\scriptsize \impmark{37}} & 0.097\rlap{\scriptsize \impmark{52}} & 0.930\rlap{\scriptsize \impmarkup{6}} & 0.794 &
0.451\rlap{\scriptsize \impmark{27}} & 0.151\rlap{\scriptsize \impmark{37}} & 0.897\rlap{\scriptsize \impmarkup{3}} & 0.775\\
\hdashline

LongLive &
0.511 & 0.218 & 0.874 & 0.817 &
0.534 & 0.170 & 0.896 & 0.806\\
\quad + Correspondence &
0.272\rlap{\scriptsize \impmark{47}} & 0.057\rlap{\scriptsize \impmark{74}} & 0.953\rlap{\scriptsize \impmarkup{9}} & 0.807 &
0.345\rlap{\scriptsize \impmark{35}} & 0.064\rlap{\scriptsize \impmark{63}} & 0.945\rlap{\scriptsize \impmarkup{5}} & 0.786\\
\hdashline

Rolling Forcing &
0.427 & 0.172 & 0.905 & 0.799 &
0.445 & 0.163 & 0.907 & 0.806\\
\quad + Correspondence &
0.240\rlap{\scriptsize \impmark{44}} & 0.066\rlap{\scriptsize \impmark{62}} & 0.955\rlap{\scriptsize \impmarkup{6}} & 0.828 &
0.305\rlap{\scriptsize \impmark{32}} & 0.058\rlap{\scriptsize \impmark{65}} & 0.952\rlap{\scriptsize \impmarkup{5}} & 0.821\\
\bottomrule
\end{tabular}
}

\label{tab:orthogonal}
\end{table}

\begin{figure*}[t!]
    \centering 
    \includegraphics[width=\linewidth]{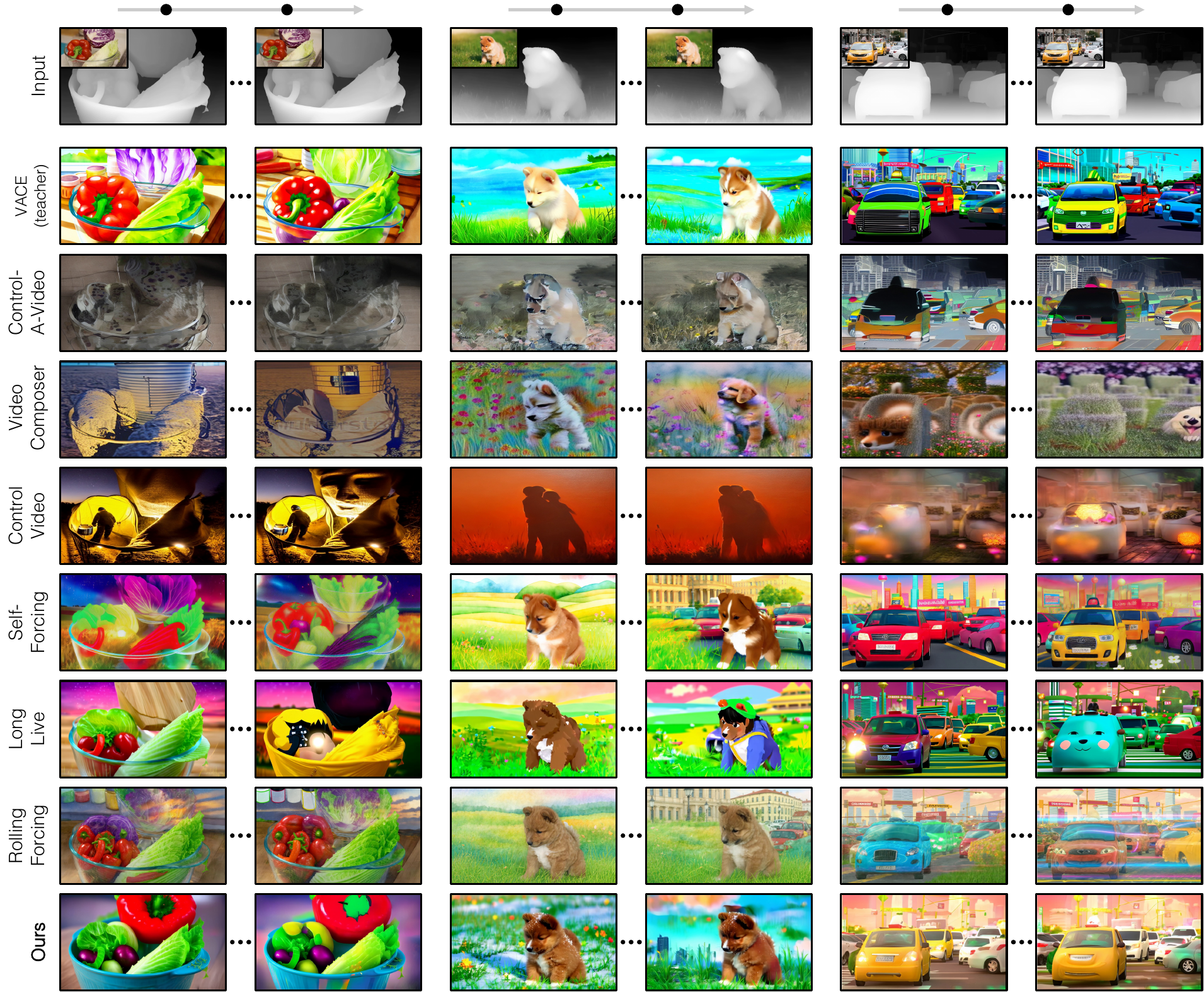}
    \caption{
    \textbf{Qualitative comparisons.} We show two frames sampled far apart in time for three example videos (left and right). Bidirectional methods (VACE 1.3B) and Autoregressive distillation methods produce high-quality individual frames but fail to maintain object identity over long durations. Adding our correspondence mechanism preserves consistent appearance throughout the video.
    }
    \label{fig:comparisons}
\end{figure*}

\begin{figure}[t!]
    \centering 
    \includegraphics[width=\linewidth]{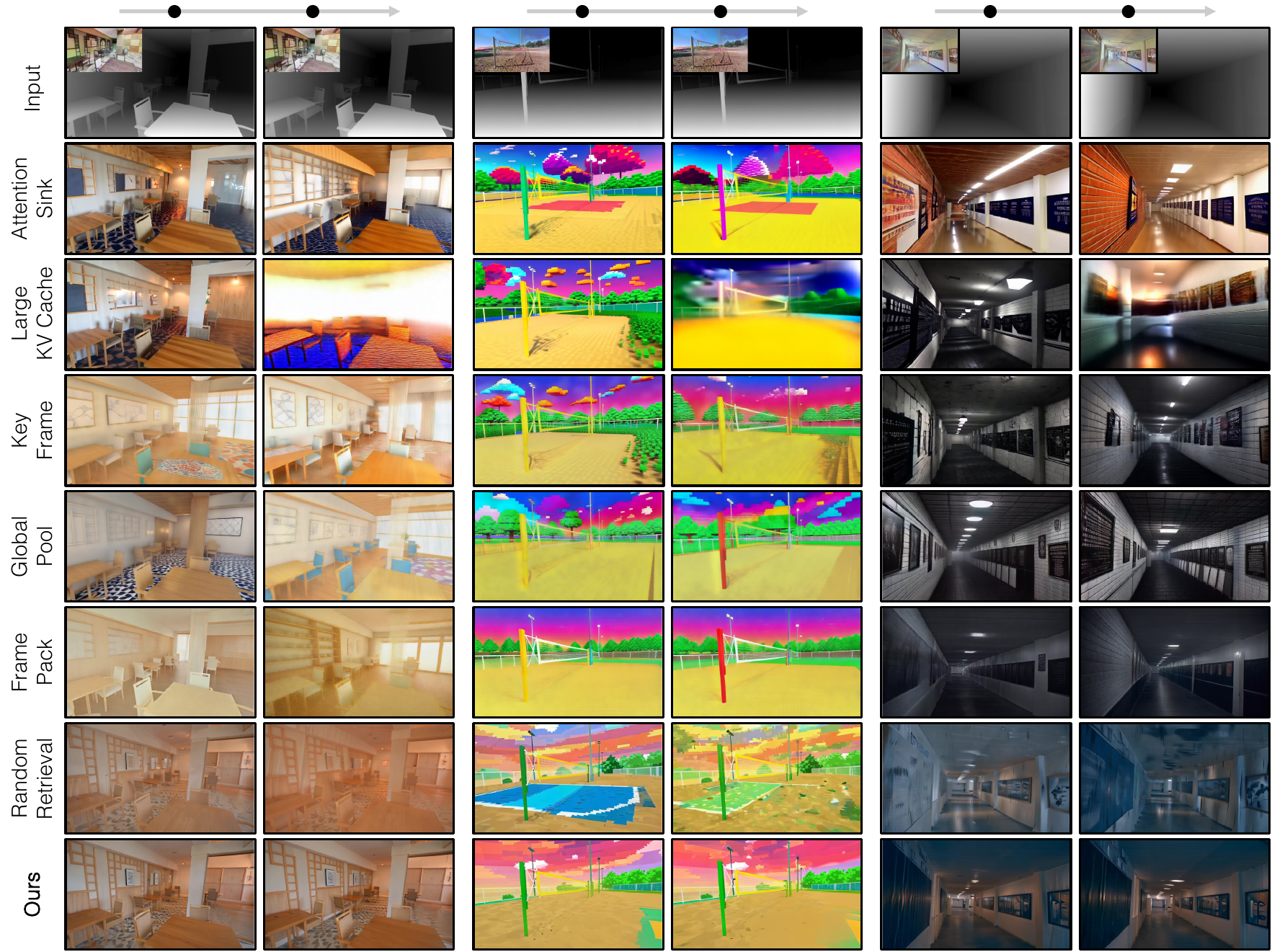}
    \caption{
    \textbf{Comparing to alternate consistency mechanisms.}
    We compare different strategies for extending the temporal context beyond sliding window, including enlarging the KV cache, keyframe retention, global pooling, FramePack~\cite{zhang2025framepack} compression, and random retrieval. 
    While these approaches provide partial improvements, they suffer for loss of details or quality degradation under long-range revisits. 
    In contrast, our correspondence based method (bottom row)  enables accurate reuse of historical content and preserving consistency over long time horizons. 
    }
    \label{fig:comparisons_alt_mem}
\end{figure}

\begin{table}[t]
\centering
\caption{\textbf{Comparison of alternate consistency mechanisms.} We compare several strategies for extending temporal context beyond the default sliding window ($W{=}21$): increasing the KV cache window size, retaining every $N$-th keyframe, maintaining a global average-pooled KV cache, Attention Sinks, and FramePack. While larger windows and keyframe retention provide modest reconstruction gains, our correspondence-based retrieval achieves substantially better long-range consistency.}
\resizebox{\textwidth}{!}{
\begin{tabular}{lcccccccc}
\toprule
& \multicolumn{4}{c}{\textbf{VACE-Bench}} & \multicolumn{4}{c}{\textbf{DL3DV}} \\
\cmidrule(lr){2-5} \cmidrule(lr){6-9}
\textbf{Method} & \textbf{LPIPS}$_\downarrow$ & \textbf{DreamSim}$_\downarrow$ & \textbf{CLIP-Sim}$_\uparrow$ & \textbf{Quality}$_\uparrow$
& \textbf{LPIPS}$_\downarrow$ & \textbf{DreamSim}$_\downarrow$ & \textbf{CLIP-Sim}$_\uparrow$ & \textbf{Quality}$_\uparrow$ \\
\midrule
KV cache ($2\times$) &
0.350 & 0.139 & 0.910 & 0.795 &
0.710 & 0.364 & 0.814 & 0.759\\
KV cache ($3\times$) &
0.308 & 0.120 & 0.920 & 0.793 &
0.770 & 0.496 & 0.744 & 0.752\\
KV cache ($4\times$) &
0.330 & 0.134 & 0.913 & 0.795 &
0.759 & 0.543 & 0.714 & 0.753\\
\hdashline
Keyframe (every 3rd) &
0.329 & 0.115 & 0.923 & 0.790 &
0.607 & 0.257 & 0.857 & 0.776\\
Keyframe (every 5th) &
0.350 & 0.128 & 0.914 & 0.789 &
0.581 & 0.240 & 0.857 & 0.776\\
Keyframe (every 10th) &
0.373 & 0.134 & 0.912 & 0.789 &
0.523 & 0.190 & 0.880 & 0.777\\
\hdashline
Global avg.\ pool &
0.450 & 0.176 & 0.891 & 0.753 &
0.457 & 0.145 & 0.909 & 0.772\\
\hdashline
Attention Sink &
0.436 & 0.159 & 0.899 & 0.793 &
0.502 & 0.156 & 0.901 & 0.788\\

\hdashline
FramePack &
\underline{0.304} & \underline{0.095} & \underline{0.938} & 0.775 &
\underline{0.344} & \underline{0.105} & \underline{0.932} & \underline{0.802}\\
\hdashline
Random retrieval &
0.396 & 0.144 & 0.916 & \underline{0.807} &
0.451 & 0.156 & 0.911 & 0.801\\
\hdashline
Correspondence (Ours) &
\textbf{0.240} & \textbf{0.066} & \textbf{0.955} & \textbf{0.828} &
\textbf{0.305} & \textbf{0.058} & \textbf{0.952} & \textbf{0.821}\\
\bottomrule
\end{tabular}
}
\label{tab:alt_consistency}
\end{table}

\begin{table}[t!]
\centering
\caption{\textbf{Importance of correspondence patches.} We measure the effect of removing the correspondence patches. Without correspondence, reconstruction metrics degrade substantially while quality scores remain similar, demonstrating the effectiveness of correspondence tokens for maintaining long term consistency.}
\resizebox{\textwidth}{!}{
\begin{tabular}{lcccccccc}
\toprule
& \multicolumn{4}{c}{\textbf{VACE-Bench}} & \multicolumn{4}{c}{\textbf{DL3DV}} \\
\cmidrule(lr){2-5} \cmidrule(lr){6-9}
\textbf{Method} & \textbf{LPIPS}$_\downarrow$ & \textbf{DreamSim}$_\downarrow$ & \textbf{CLIP-Sim}$_\uparrow$ & \textbf{Quality}$_\uparrow$
& \textbf{LPIPS}$_\downarrow$ & \textbf{DreamSim}$_\downarrow$ & \textbf{CLIP-Sim}$_\uparrow$ & \textbf{Quality}$_\uparrow$ \\
\midrule
Ours w/o Corr. &
0.379 & 0.160 & 0.910 & 0.829 &
0.437 & 0.124 & 0.921 & 0.822\\
Ours &
0.240\rlap{\scriptsize \impmark{37}} & 0.066\rlap{\scriptsize \impmark{59}} & 0.955\rlap{\scriptsize \impmarkup{5}} & 0.828 &
0.305\rlap{\scriptsize \impmark{30}} & 0.058\rlap{\scriptsize \impmark{53}} & 0.952\rlap{\scriptsize \impmarkup{3}} & 0.821\\
\bottomrule
\end{tabular}
}
\label{tab:abl_corr}
\end{table}

\section{Experiments}
\label{sec:experiments}
Here, we first evaluate our method on depth-to-video translation, focusing on long-range temporal consistency on long videos. Our mechanism is agnostic to the conditioning modality. 
Experiments are conducted on two benchmarks: VACE-Benchmark~\cite{vace}, covering diverse real-world scenes, and DL3DV~\cite{ling2024dl3dv}, a geometry-grounded 3D dataset enabling controlled pose revisits.
We compare against bidirectional diffusion models, prior video translation methods, and recent autoregressive distillation approaches, and ablate the key components of our correspondence mechanism.
Unless stated otherwise, all experiments use VACE 1.3B at 480p resolution.

\myparagraph{Datasets.}
We evaluate on two benchmarks. 
VACE-Benchmark~\cite{vace} contains diverse real-world video sequences with varying motion and scene complexity.
DL3DV~\cite{ling2024dl3dv} is a geometry-grounded 3D dataset with calibrated camera poses, enabling controlled evaluation of viewpoint revisits and long-range consistency.
Additional dataset details and preprocessing are provided in appendix.

\myparagraph{Evaluation Protocol.}
We evaluate our method in terms of video quality, long-range consistency under revisits, and inference efficiency.
Video quality is measured on full generated videos using V-Bench~\cite{huang2023vbench} metrics (subject consistency, background consistency, temporal flickering, motion smoothness, aesthetic quality, and imaging quality). Experiments report average of scores, individual metrics values are provided in the appendix. 
Long-range consistency is evaluated using controlled revisit protocols that require model to reproduce previously observed content.
On VACE-Benchmark, following~\cite{guo2026efficient}, we construct synthetic long sequences by concatenating two distinct 5-second clips in an interleaved order (A→B→A→B), and measure reconstruction consistency between repeated occurrences of the same clip.
Consistency is quantified using LPIPS~\cite{zhang2018perceptual}, DreamSim~\cite{fu2023dreamsim}, and CLIP~\cite{radford2021learning} image similarity.
On DL3DV, we use pretrained 3D Gaussian Splatting models~\cite{kerbl3Dgaussians} using calibrated camera poses to render exact pose revisits and measure consistency between first-visit and revisit frames.
We measure inference latency and throughput. 
Inference latency is measured as time to produce first frame and throughput is reported in frames per second (FPS). All measurements are performed on a single NVIDIA H100 GPU.

\myparagraph{Implementation Details.}
We use VACE 1.3B~\cite{vace} as the student model and VACE 14B as the teacher.
The student is initialized from the pretrained VACE 1.3B checkpoint and trained using Rolling Forcing~\cite{liu2025rolling}.
For correspondence retrieval, we use VAE encoder's latent features as keys, with $k = 3$ retrieved neighbors, history threshold $\tau_h = 0.75$, and memory threshold $\tau_m = 0.75$.
Training is performed using AdamW~\cite{loshchilov2017decoupled} with batch size 64, a learning rate of $2\times10^{-6}$ for the student network and $4\times10^{-7}$ for the critic.

\subsection{Baseline Comparisons}
\label{sec:baseline_comparison}

We evaluate our method against several video-to-video translation frameworks. Quantitative and qualitative results are in Table~\ref{tab:memory_rec} and Figure~\ref{fig:comparisons}.
First, we compare to 
Control-A-Video~\cite{chen2023control}, VideoComposer~\cite{wang2023videocomposer}, and ControlVideo~\cite{zhang2023controlvideo}.

They operate in an offline bidirectional setting with multi-step diffusion and global attention over fixed-length sequences.
In our experiments, they produce lower visual quality and struggle to maintain long-range consistency (Figure~\ref{fig:comparisons}).
ControlVideo achieves relatively strong reconstruction scores due to its hierarchical sampling strategy over long sequences.
However, this comes at the cost of high latency ($\approx$ 42 minutes in our setting) and reduced perceptual quality, making it impractical for real-time video-to-video translation.

We next compare to newer, state-of-the-art video translation models, such as VACE~\cite{vace}.
As discussed before, VACE is a bidirectional model which is limited to a fixed-length, short video sequences. 
Unfortunately, there is no publicly available adaption of VACE to an unbounded-length, streaming use case. 
In order to still apply VACE on our setting, we adapt it in two manners.

First, we directly evaluate VACE by applying it on non-overlapping video segments. 
We find that this naive adaptation of VACE produces high-quality results, but unsurprisingly has poor long-range consistency. 
For example, note row~2 in Figure~\ref{fig:comparisons} where each individual frame is of high-quality, but objects change their colors drastically over the length of the video.

Second, we adapt VACE to a streaming use-case with three recent causal, autoregressive distillation methods: Self-Forcing~\cite{huang2025self}, LongLive~\cite{yang2025longlive}, and Rolling Forcing~\cite{liu2025rolling}. 
These methods were originally designed for text-to-video generation and aren't applied to video-to-video yet. 
Here, we implement and adapt the three methods to video-to-video with VACE.
We find that all methods achieve competitive quality scores and substantially lower latency than the bidirectional VACE baseline (Tables~\ref{tab:memory_rec},~\ref{tab:memory_rec_dl3dv}). 
Nevertheless, these methods still perform poorly on long-range consistency.
On VACE-Bench, Self-Forcing and LongLive underperform bidirectional VACE, while Rolling Forcing only slightly outperforms it in long-range reconstruction.

Our main contribution, correspondence mechanism, can be readily applied with any autoregressive method. In Table~\ref{tab:memory_rec},~\ref{tab:memory_rec_dl3dv} and Figure~\ref{fig:comparisons}, we report results of our best configuration, where our method is applied with Rolling Forcing. 
Among autoregressive methods, our approach achieves the strongest long-range consistency and visual quality while maintaining real-time performance.
In Table~\ref{tab:orthogonal}, we report the results of our method when applied on top of all autoregressive methods. 
Adding our correspondence mechanism consistently improves long-term consistency across all autoregressive backbones while preserving visual quality. Table~\ref{tab:orthogonal} results demonstrate that our method is orthogonal to choice of autoregressive backbone. 
Please refer to the appendix for a comparison of different conditioning mechanisms.

\begin{figure}[t!]
    \centering 
    \includegraphics[width=\linewidth]{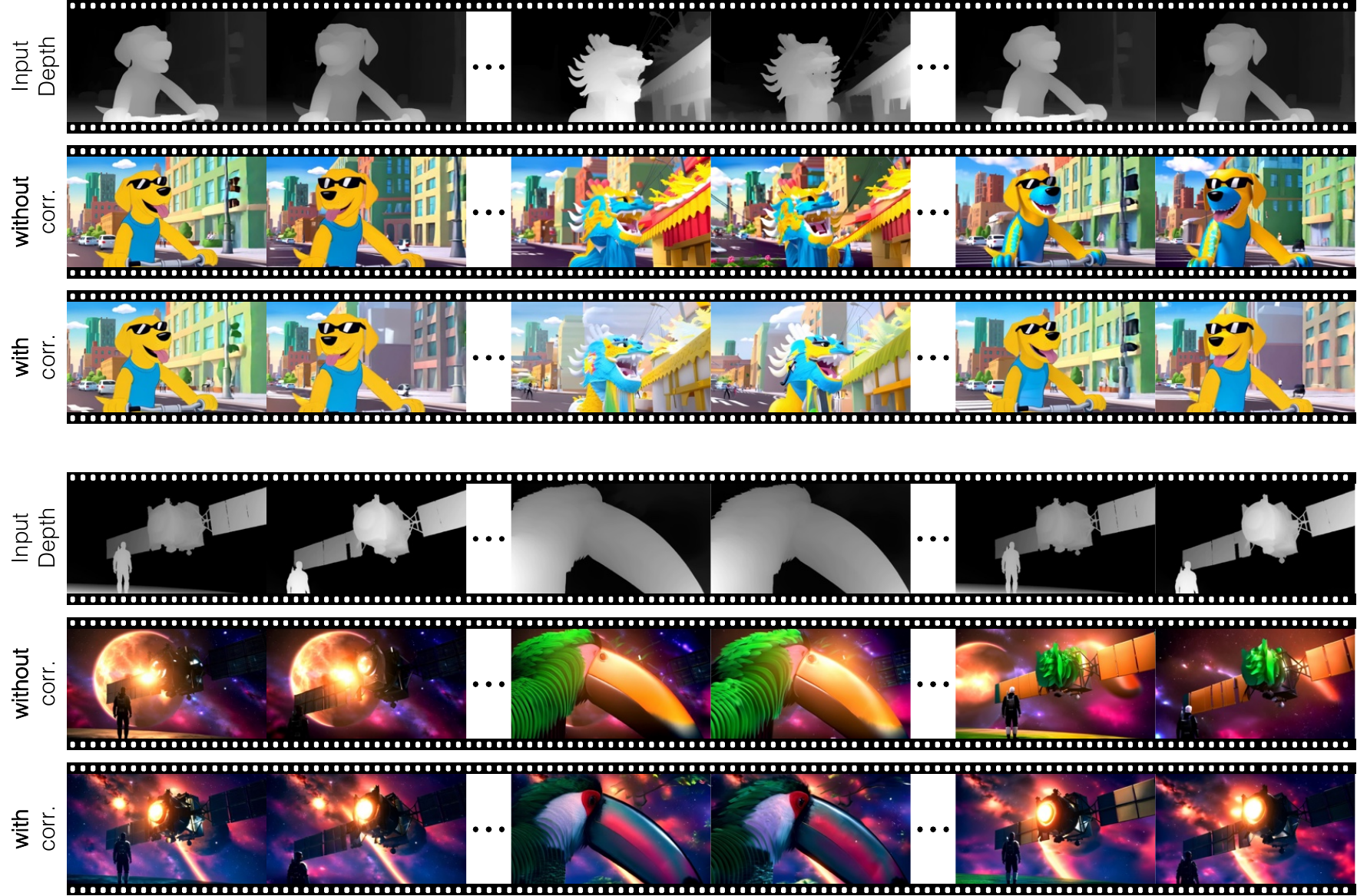}
    \caption{
    \textbf{Effect of correspondence pointers on long-range consistency.} Without correspondence pointers, the model generates plausible but inconsistent frames. Object appearance drifts over time. Consider  the top example, the dog starts off having a yellow snout, but it changes to blue  when it re-appears. 
    With correspondence pointers, the model retrieves and reuses visual context from earlier frames, maintaining consistent identity throughout the video. 
    For example in the top video, the dog snout is yellow at the beginning of the video. However it turns into a blue snout when reappearing if correspondence is not used. However it stays consistently yellow with our method. 
    }
    \label{fig:abl_correspondence}
\end{figure}
\vspace{-4mm}

\subsection{Alternate Consistency Mechanisms}
\label{sec:consistency_comparison}
We compare our correspondence mechanism against alternative strategies for extending temporal context beyond the default sliding window.
Qualitative and quantitative results are shown in Figure~\ref{fig:comparisons_alt_mem} and Table~\ref{tab:alt_consistency}, respectively.

\myparagraph{Larger KV Cache window.}
We increase the sliding window size from 21 latent frames to 42 ($\times2$), 63 ($\times3$), and 84 ($\times4$).
 On VACE-Bench, larger windows improve reconstruction consistency slightly, although the gains are non-monotonic. 
However, on DL3DV, where sequences are substantially longer, performance degrades sharply in both quality and long-range consistency.

\myparagraph{Keyframe retention.}
Next, instead of sliding window, we retain KV cache of every $N$-th block ($N{=}3,5,10$), providing a sparse but more complete coverage of the history.
While this improves over simply enlarging the window, it remains worse than correspondence-based retrieval in long-range reconstruction (Table~\ref{tab:alt_consistency}).

\myparagraph{Global Pooling.}
We maintain a single block of KV tokens as a running cumulative average of all previous blocks, providing constant memory access to the entire history.
Although this improves long-range consistency on DL3DV degrades visual quality on both datasets (quality scores drop to 0.753 and 0.772).

\myparagraph{Random Retrieval.}
To isolate the role of correspondence matching from the cross-attention layers, we replace the similarity-based retrieval with uniform random sampling from the memory bank.
Each query patch receives $k$ randomly selected history patches instead of the top-$k$ most similar, the history creation process remains the same. 
This significantly degrades long-range reconstruction performance, demonstrating that accurate correspondence matching is essential.

\myparagraph{Attention Sink.}
Recent methods such as LongLive, MotionStream, and Rolling Forcing retain initial frames as an attention sink to provide global context.
However, this mechanism cannot capture novel content that appears later in the sequence and is insufficient to prevent identity drift in long videos.

\myparagraph{FramePack.}
FramePack~\cite{zhang2025framepack} is applied to the task of video to video translation, storing past blocks at geometrically decreasing spatial resolution. Recent blocks maintain full resolution while older blocks are progressively downsampled. 
While this retains the entire history, aggressive spatial compression reduces detail, leading to weaker long-range reconstruction.

Overall, correspondence-based retrieval outperforms all alternative mechanisms for extending temporal context, achieving stronger long-range consistency without sacrificing visual quality.

\subsection{Analysis and Ablations}
\label{sec:analysis}
\myparagraph{Importance of Correspondence.}
First, we show the importance of using correspondence patches through an ablation where we remove the correspondences and disable the correspondence cross-attention layer (Equation~\ref{eq:corr_xa}).  
As in Table~\ref{tab:abl_corr}, without using correspondence the long context consistency gets significantly worse. The LPIPS reconstruction improves by $37\%$ and $30\%$ on the two datasets. DreamSim and CLIP similarity scores are also similarly effected.
Figure~\ref{fig:abl_correspondence} shows this qualitatively. 
The top row shows the generated video when the correspondence is disabled. In this video, the dog starts with a yellow snout, but as the video progresses and the dog re-enters the scene, the snout turns blue.
In contrast, our full method with correspondence is shown in the bottom row. Here the dog retains the identity and re-enters the scene with the same yellow snout. 

\myparagraph{Sensitivity to different distillation methods.}
Next, we show that our method is not specific to any single distillation method. We apply our method to three different distillation methods: SelfForcing~\cite{huang2025self}, LongLive~\cite{yang2025longlive}, and RollingForcing~\cite{liu2025rolling}. 
Add our correspondence method improves the consistency across each of the base distillation methods as seen in Table~\ref{tab:orthogonal}.

\section{Conclusion and Limitations}
\label{sec:conclusion}

We introduce \texttt{vid2vid-long}, a correspondence-based memory mechanism for long-range consistent video-to-video translation. By exploiting a structural property of the task, matching regions in the input should match in the output, we replace implicit long-range attention with explicit input-driven retrieval.
Our dynamic memory grows with visual novelty rather than length of the video, enabling sub-linear scaling while preserving real-time performance. 
Across multiple autoregressive distillation backbones, our method consistently improves long-range reconstruction and reduces identity drift while maintaining high visual quality and real-time inference performance.
Compared to alternative context extension strategies, explicit correspondence retrieval proves both more effective and more efficient.

\myparagraph{Limitations.}
Our model struggles on fast videos with large motion, where reliable cross-frame correspondences become harder to establish. It can also fail when the text prompt is inconsistent with the input depth video, since the model must reconcile conflicting semantic and structural conditioning signals.

\myparagraph{Acknowledgment.}
We thank Sheng-Yu Wang, Nupur Kumari, Maxwell Jones, and Xun Huang for their helpful comments and discussion. The project was partially done during an internship with Adobe Research. The project is partially funded by NSF IIS-2239076, NSF ISS-2403303, the IITP grant funded by the Korean Government (MSIT) (No. RS-2024-00457882, National AI Research Lab Project), and
the Packard Fellowship.

\clearpage

\clearpage
\setcounter{section}{0}
\renewcommand\thesection{\Alph{section}}

\begin{center}
  \Large\textbf{Appendix}
\end{center}
\vspace{1em}

We provide additional details and results that complement the main paper.
Section~\ref{sec:supp_architecture} presents the architecture diagram and runtime analysis.
Section~\ref{sec:supp_history_growth} analyzes the history growth and memory scaling behavior.
Section~\ref{sec:supp_other_conditions} shows results on other conditioning inputs. 
Sections~\ref{sec:supp_training} and~\ref{sec:supp_inference} provide training and inference details.
Section~\ref{sec:supp_feature_extractor} compares different feature extractors for correspondence matching.
Section~\ref{sec:supp_dataset} details the dataset preprocessing steps.
Section~\ref{sec:supp_vbench} reports per-metric V-Bench scores.
The video results are primarily shown in the accompanying website.

\section{Model Architecture and Runtime Analysis}
\label{sec:supp_architecture}
\begin{figure}[t!]
    \centering 
    \includegraphics[width=\linewidth]{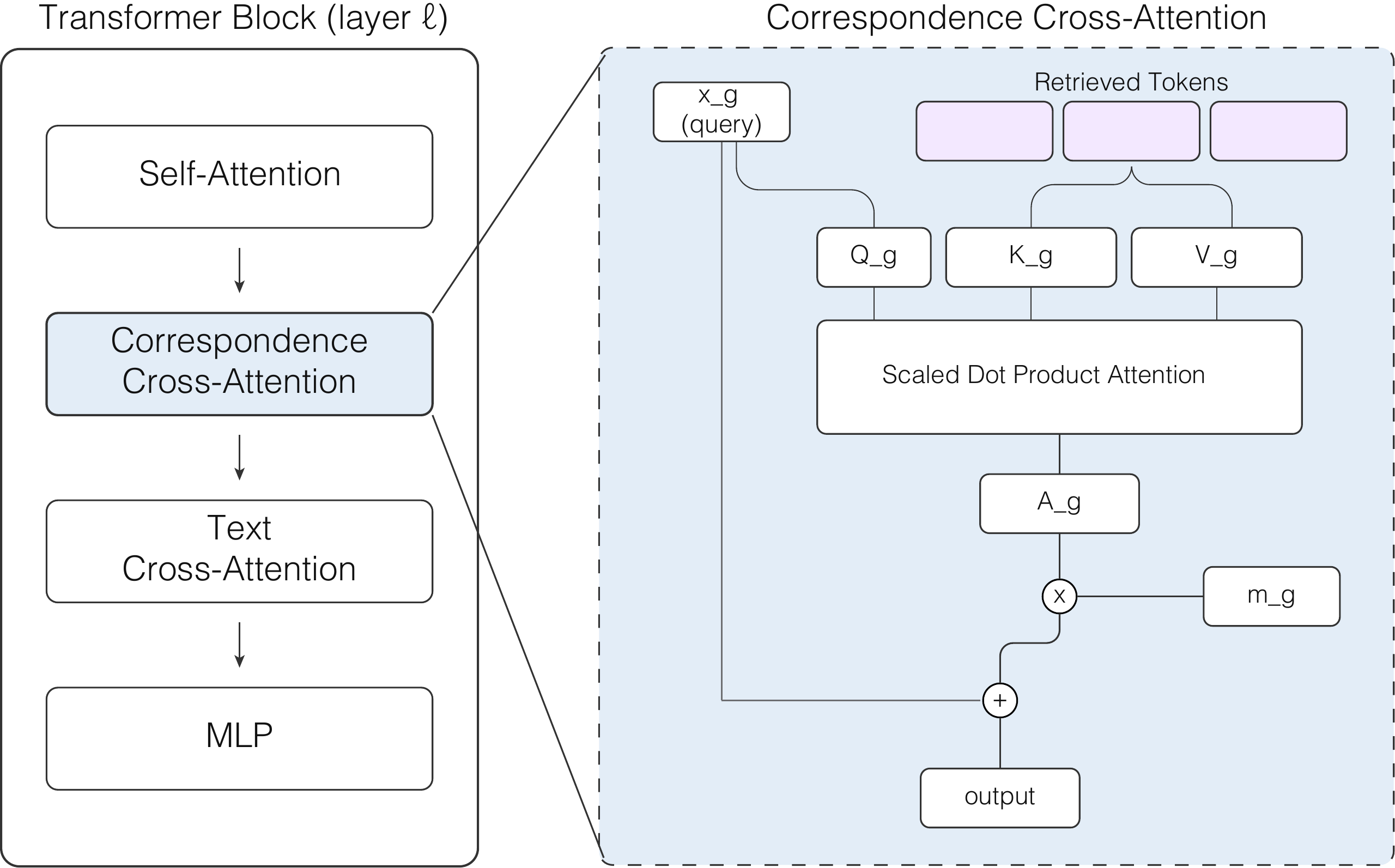}
    \caption{
    \textbf{Network architecture.}
    We show a single transformer block at layer $\ell$ on the left. Our method inserts a new correspondence cross-attention layer between existing self-attention and text cross-attention layers. Details of the correspondence cross-attention are shown on the right. For each token at index $g$, the current hidden state $x_g$ serves as the query and is projected to query $Q_g$. The $k$ retrieved neighbors are projected to keys $K_g$ and values $V_g$ (see Section 3.3 of main paper). The standard scaled dot-product attention produces the output $A_g$, which is projected through $W_o$ and multiplied by the binary mask $m_g$. The result is added back to $x_g$ as a residual connection. 
    }
    \label{fig:arch_corrxa}
\end{figure}

Section~\ref{sec:correspondence} of the main paper describes conditioning with correspondence-cross attention layers. In Figure~\ref{fig:arch_corrxa}, we show the corresponding architecture diagram.
The only architectural modification to
the pretrained VACE transformer is the insertion of a correspondence
cross-attention layer in each block. Note that each token attends only to its
$k$ retrieved neighbors.

\begin{figure}[t!]
    \centering
    \includegraphics[width=\linewidth]{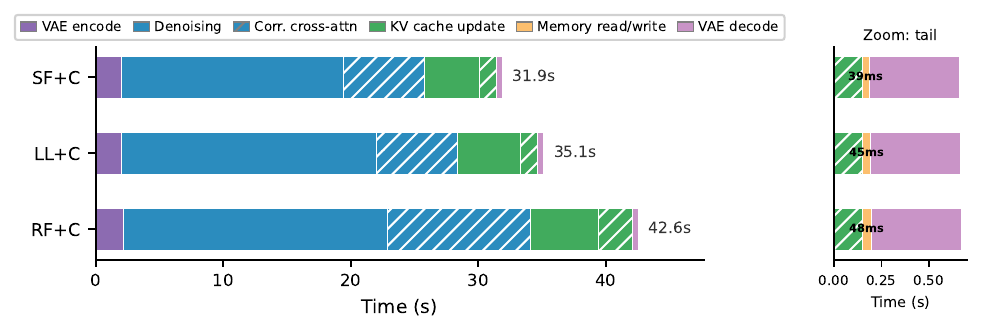}
    \vspace{-6mm}
    \caption{
    \textbf{Per-component runtime breakdown.}
    Inference time for generating a 20-second video (320 frames) on a single H100 GPU at 480p, for each autoregressive backbone with our correspondence mechanism.
    The denoising forward passes dominate the total runtime across all methods.
    Hatched regions indicate the correspondence cross-attention overhead within the denoising and KV cache components.
    The inset zooms into the memory read/write component, which is negligible (${<50}$ms) across all methods.
    }
    \label{fig:runtime_breakdown}
    \vspace{-4mm}
\end{figure}

Figure~\ref{fig:runtime_breakdown} reports the per-component inference time for generating a 20-second video on a single H100 GPU at 480p resolution, for all three autoregressive backbones with our correspondence mechanism.
The denoising forward passes dominate the total runtime across all methods, accounting for over 77\% of the total.
The correspondence-specific components (memory read, correspondence cross-attention, and memory write) add a small overhead. Memory read and write operations have a negligible inference cost (${\sim}50$ms total).

\section{History Growth Analysis}
\label{sec:supp_history_growth}

As discussed in Section~\ref{sec:method} of the main paper, the memory footprint scales with the visual novelty of the scene rather than the raw sequence length. Here, we empirically validate this claim.

\myparagraph{VRAM Scaling.}
Table~\ref{tab:vram_scaling} compares the total GPU memory footprint of a naive full KV-cache approach against our dynamic memory mechanism as video length increases.
The naive approach retains KV entries for every past token across all 45 transformer layers, causing memory to grow linearly with sequence length. At 480p resolution, it exceeds the 80\,GB capacity of an H100 GPU after roughly 30 seconds; at 720p the limit is reached even sooner.
Our method replaces the large KV cache with an attention sink and a fixed-size sliding window ($W{=}21$ latent frames), plus a correspondence memory that stores only input--output latent patch pairs.
Because the correspondence memory bypasses the per-layer KV overhead, its footprint is negligible: even after 10 minutes of video (9,600 frames), total VRAM increases by only 0.1\,GB at 480p and 0.3\,GB at 720p compared to the initial cost.
As a result, total VRAM remains nearly constant regardless of video duration at both resolutions.

\begin{table}[t!]
\centering
\caption{\textbf{VRAM scaling with video length.}
Full KV-cache retains all past tokens across all transformer layers, causing memory to grow linearly with video length.
Our method replaces the unbounded cache with a fixed kv cache plus a lightweight correspondence memory.
Total VRAM remains nearly constant regardless of video duration at both resolutions.
OOM denotes exceeding the 80\,GB capacity of an H100 GPU.
}
\label{tab:vram_scaling}
\setlength{\tabcolsep}{5pt}
\begin{tabular}{rc cc cc c}
\toprule
& & \multicolumn{2}{c}{\textbf{480p (480$\times$832)}} & \multicolumn{2}{c}{\textbf{720p (720$\times$1280)}} & \\
\cmidrule(lr){3-4} \cmidrule(lr){5-6}
\textbf{Duration} & \textbf{Frames} & \textbf{Full KV} & \textbf{Ours} & \textbf{Full KV} & \textbf{Ours} & $|\mathcal{H}|$ \\
\midrule
10\,s & 160 & 42.2 & 34.1 & 63.7 & 45.2 & 2\,K \\
30\,s & 480 & 74.3 & 34.1 & OOM & 45.2 & 4\,K \\
1\,min & 960 & OOM & 34.1 & OOM & 45.2 & 6\,K \\
2\,min & 1920 & OOM & 34.1 & OOM & 45.3 & 9\,K \\
5\,min & 4800 & OOM & 34.2 & OOM & 45.4 & 16\,K \\
10\,min & 9600 & OOM & 34.2 & OOM & 45.5 & 26\,K \\
\bottomrule
\end{tabular}
\end{table}

\begin{figure}[t!]
    \centering
    \includegraphics[width=\linewidth]{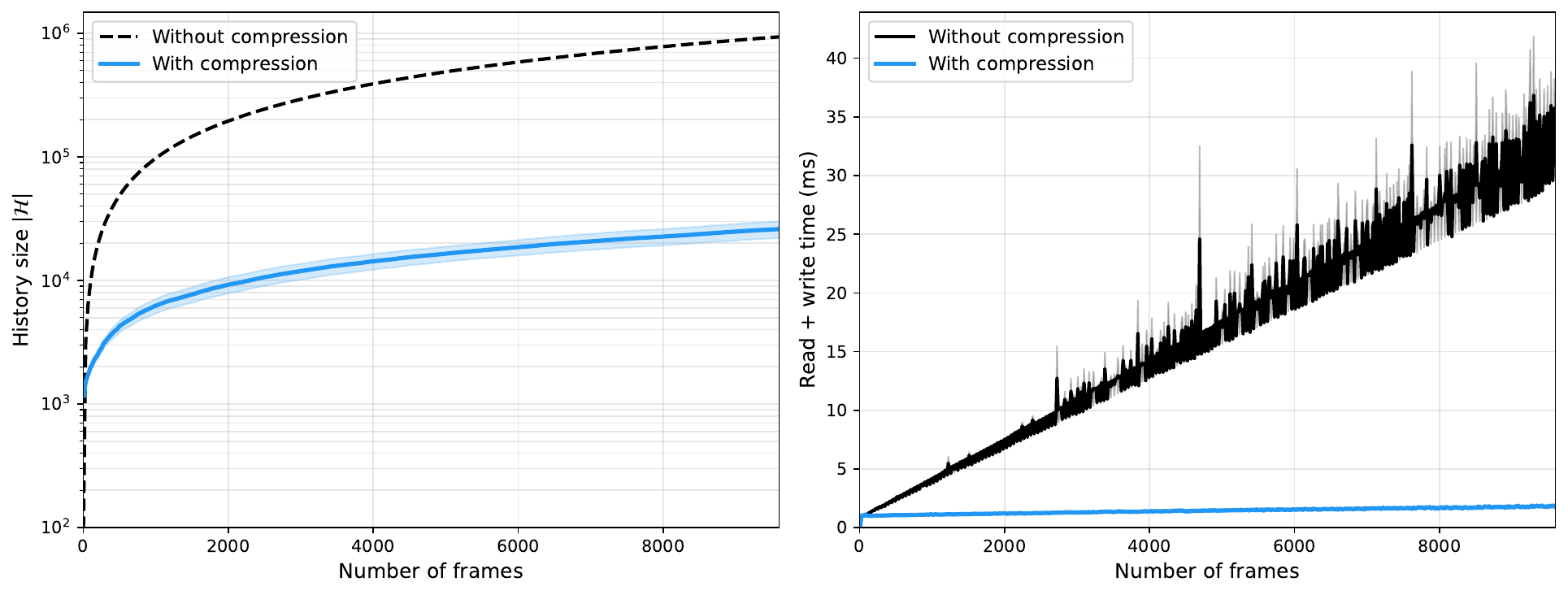}
    \vspace{-6mm}
    \caption{
    \textbf{Effect of history compression.}
    \textbf{Left:} History size $|\mathcal{H}|$ vs.\ video length (log scale). Without compression, the history grows linearly to over 936\,K entries at 9\,600 frames (10 minutes). With our redundancy-based write policy ($\tau_h{=}0.75$), the history grows sub-linearly to ${\sim}$26\,K entries (${\sim}36\times$ reduction).
    \textbf{Right:} Corresponding per-block memory read and write time. Without compression, the cost grows linearly, exceeding 30\,ms at 10 minutes. With compression, it remains below 2\,ms throughout. Shaded regions show $\pm$1 standard error across 30 videos spanning three categories (webcam, surveillance, action camera).
    }
    \label{fig:history_growth}
    \vspace{-4mm}
\end{figure}

\myparagraph{History Growth and Read/Write Cost.}
Figure~\ref{fig:history_growth} (left) shows the history size $|\mathcal{H}|$ over time with and without our novelty-based compression, evaluated on 30 videos spanning three categories (webcam, surveillance, and action camera).
Without compression, every generated patch is stored, causing $|\mathcal{H}|$ to grow linearly to over 936\,K entries at 9\,600 frames (10 minutes).
With our write policy ($\tau_h{=}0.75$), the history grows sub-linearly, reaching roughly 26\,K entries at the same point (${\sim}36\times$ reduction).
This directly translates to runtime savings: Figure~\ref{fig:history_growth} (right) shows that the per-block memory read and write time grows linearly without compression, exceeding 30\,ms at 10 minutes, while with compression it remains below 2\,ms throughout.

\myparagraph{History Growth and thresholds $\tau_h$ and $\tau_m$}
Increasing $\tau_h$ admits more patches into the memory, trading memory size for better reconstruction. 
LPIPS improves from $0.301$ to $0.204$ as $|\mathcal{H}|$ grows from $413$ to $15.8K$ patches.
For $\tau_m$, the reconstruction quality is stable over a wide range of values. 
But, a strict threshold ($\tau_m = 0.90$) rejects many valid matches, degrading reconstruction (LPIPS $0.286$).
The same defaults perform consistently for both datasets with no scene-specific tuning needed.

\begin{table}[t!]
\centering
\caption{\textbf{Hyperparameter analysis.}
We show the effects of varying the thresholds $\tau_h$ and $\tau_m$ on our method.
As the threshold $\tau_h$ is increased, our method stores more patches in the history bank $\mathcal{H}$ and improves the reconstruction performance.
}
\label{tab:tau_hm_effects}
\setlength{\tabcolsep}{5pt}
\centering\footnotesize
\begin{minipage}{0.54\linewidth}
\centering
\begin{tabular}{ccccc}
\toprule
$\tau_h$ & $|\mathcal{H}|$ & LPIPS$\downarrow$ & CLIP$\uparrow$ & DreamSim$\downarrow$ \\
\midrule
0.25 & 413      & 0.301 & 0.940 & 0.092 \\
0.50 & 682      & 0.284 & 0.946 & 0.081 \\
0.75 & 3{,}897  & 0.256 & 0.957 & 0.060 \\
0.90 & 15{,}822 & 0.204 & 0.963 & 0.049 \\
\bottomrule
\end{tabular}
\end{minipage}%
\hspace{3pt}\vrule\hspace{3pt}%
\begin{minipage}{0.43\linewidth}
\centering
\begin{tabular}{cccc}
\toprule
$\tau_m$ & LPIPS$\downarrow$ & CLIP$\uparrow$ & DreamSim$\downarrow$ \\
\midrule
0.25 & 0.248 & 0.956 & 0.059 \\
0.50 & 0.249 & 0.957 & 0.058 \\
0.75 & 0.256 & 0.957 & 0.060 \\
0.90 & 0.286 & 0.946 & 0.084 \\
\bottomrule
\end{tabular}
\end{minipage}

\end{table}

\section{Other Conditioning Inputs}
\label{sec:supp_other_conditions}
We show the evaluation of different methods when other conditioning inputs such as sketch and optical flow are used in Table~\ref{tab:other_conditions}. Our method is robust and shows an improvement in the long range consistency across other conditioning inputs (sketch and optical flow) too. 

\begin{table}[t!]
\centering
\caption{\textbf{Other conditioning inputs}
We show evaluation with sketch inputs (left) and optical flow inputs (right). 
}
\label{tab:other_conditions}
\setlength{\tabcolsep}{5pt}
\begin{minipage}{0.47\linewidth}
\centering
\resizebox{\linewidth}{!}{
\begin{tabular}{lccc}
\toprule
Method & LPIPS$\downarrow$ & CLIP$\uparrow$ & DreamSim$\downarrow$ \\
\midrule
Self-Forcing    & 0.474 & 0.906 & 0.167 \\
LongLive        & 0.384 & 0.911 & 0.156 \\
Rolling Forcing & 0.284 & 0.940 & 0.100 \\
Ours            & \textbf{0.191} & \textbf{0.960} & \textbf{0.050} \\
\bottomrule
\end{tabular}
}
\end{minipage}%
\hspace{2pt}\vrule\hspace{2pt}%
\begin{minipage}{0.47\linewidth}
\centering
\resizebox{\linewidth}{!}{
\begin{tabular}{lccc}
\toprule
Method & LPIPS$\downarrow$ & CLIP$\uparrow$ & DreamSim$\downarrow$ \\
\midrule
Self-Forcing    & 0.535 & 0.852 & 0.253 \\
LongLive        & 0.516 & 0.858 & 0.235 \\
Rolling Forcing & 0.494 & 0.868 & 0.226 \\
Ours            & \textbf{0.263} & \textbf{0.934} & \textbf{0.092} \\
\bottomrule
\end{tabular}
}
\end{minipage}
\end{table}

\section{Training Details}
\label{sec:supp_training}

Table~\ref{tab:supp_hyperparams} lists all training and model hyperparameters.
We train the student model using Rolling Forcing~\cite{liu2025rolling} distillation with 5 uniformly spaced denoising steps ($\{1000, 800, 600, 400, 200\}$) that are shifted following standard practice.
The student is initialized from the pretrained VACE 1.3B checkpoint and all parameters are finetuned jointly, including the newly introduced correspondence cross-attention layers.
The student is initialized with Diffusion Forcing training before timestep distillation.
The teacher is the bidirectional VACE 14B model and does not use correspondence memory.
During training, the correspondence memory is active: for each rollout, memory read and write operations are performed as described in Section~\ref{sec:history} of the main paper.
We use the standard Rolling Forcing objective without modification or additional loss terms.
We apply EMA with weight 0.99 starting at iteration 200. Training uses mixed precision and gradient checkpointing on 32 A100 80GB GPUs across 4 nodes. All inference benchmarks are measured on a single H100 80GB GPU.

\begin{table}[h]
\centering
\caption{Training and model hyperparameters.}
\label{tab:supp_hyperparams}
\setlength{\tabcolsep}{10pt}
\begin{tabular}{@{}lc@{}}
\toprule
\textbf{Hyperparameter} & \textbf{Value} \\
\midrule
Student model & VACE 1.3B \\
Teacher model & VACE 14B \\
Distillation method & Rolling Forcing \\
Distillation loss & DMD \\
Denoising steps & 5 \\
Guidance scale & 5.0 \\
\hdashline\noalign{\vskip 3pt}
Patch descriptor $\phi$ & VAE latent \\
Retrieved neighbors $k$ & 3 \\
History threshold $\tau_h$ & 0.75 \\
Matching threshold $\tau_m$ & 0.75 \\
Per-patch correspondence dropout & 0.1 \\
Full-frame correspondence dropout & 0.1 \\
\hdashline\noalign{\vskip 3pt}
Optimizer & AdamW \\
Student learning rate & $2{\times}10^{-6}$ \\
Critic learning rate & $4{\times}10^{-7}$ \\
$(\beta_1, \beta_2)$ & $(0.0, 0.999)$ \\
Weight decay & 0.01 \\
Batch size & 64 \\
Training iterations & 2000 \\
Training resolution & 480p \\
Training latent frames & 21 \\
Latent frames per block & 3 \\
Sliding window $W$ (latent frames) & 21 \\
\bottomrule
\end{tabular}
\end{table}

\myparagraph{Negative Prompt.} Following standard practice, we use a fixed negative prompt: \textit{``Bright tones, overexposed, static, blurred details, subtitles, style, works, paintings, images, static, overall gray, worst quality, low quality, JPEG compression residue, ugly, incomplete, extra fingers, poorly drawn hands, poorly drawn faces, deformed, disfigured, misshapen limbs, fused fingers, still picture, messy background, three legs, many people in the background, walking backwards.''}

\section{Inference Details}
\label{sec:supp_inference}

Algorithm~\ref{alg:inference} provides the complete inference procedure.
The model processes the input conditioning video block-by-block.
For each block, we retrieve matching past outputs from the correspondence memory (\textbf{memory read}), denoise with correspondence cross-attention conditioning, and update the memory with novel input-output pairs (\textbf{memory write}).
Retrieval is performed once per block and the retrieved tokens are reused across all denoising steps.

\begin{figure}[t!]
\begin{minipage}{1.0\linewidth}
\begin{algorithm}[H]
\caption{Inference of \texttt{vid2vid-long}}
\label{alg:inference}
\begin{lstlisting}
# model: Diffusion transformer w/ corr. cross-attn
# vae: Pretrained encoder $\color{codeblue}\mathcal{E}$ / decoder $\color{codeblue}\mathcal{D}$
# video: Input conditioning (e.g., depth maps)
# text: Text prompt for stylization
# tau_h, tau_m: History / matching thresholds
# k: Number of retrieved neighbors per patch
# W: Sliding window size (latent frames)
# ts: Denoising schedule [1000, 800, 600, 400, 200]

def vid2vid_long(model, vae, video, text,
                 tau_h, tau_m, k, W, ts):
    H_k, H_v = [], []          # memory $\color{codeblue}\mathcal{H}$: keys + values
    kv_cache = None

    for block in chunk(video, block_size):
        # Encode input to latent patches
        z_in = vae.encode(block)
        q = patchify(z_in)                    # [N, c]

        # --- Memory Read (Eq. 3) ---
        corr, mask = zeros(N, k, c), zeros(N)
        if len(H_k) > 0:
            S = cosine_sim(q, stack(H_k))     # [N, |$\color{codeblue}\mathcal{H}$|]
            vals, idx = topk(S, k, dim=1)
            hit = vals >= tau_m
            corr[hit] = stack(H_v)[idx[hit]]
            mask[hit] = 1.0

        # --- Denoise (corr reused across all steps) ---
        z = randn_like(z_in)
        for t in ts:
            z = denoise(z, t, z_in, text, corr, mask, kv_cache)

        # --- Memory Write (Eq. 2) ---
        v = patchify(z)
        s = max_cosine_sim(q, H_k)
        novel = s < tau_h
        H_k.extend(q[novel])
        H_v.extend(v[novel])

        # Update sliding window KV cache
        kv_cache = slide_kv(kv_cache, W)
        yield vae.decode(z)
\end{lstlisting}
\end{algorithm}
\end{minipage}
\end{figure}

\begin{table*}[t]
\centering
\caption{\textbf{Ablation of the feature extractor $\phi$.} We compare three choices for computing patch descriptors: VAE latent features, DINOv2, and CLIP. VAE latent features achieve the best reconstruction on VACE-Bench and competitive performance on DL3DV, without requiring an external feature model. All variants use $k{=}3$ retrieved neighbors.}
\label{tab:abl_phi}

\resizebox{\textwidth}{!}{%
\begin{tabular}{lcccccccc}
\toprule
& \multicolumn{4}{c}{\textbf{VACE-Bench}} & \multicolumn{4}{c}{\textbf{DL3DV}} \\
\cmidrule(lr){2-5} \cmidrule(lr){6-9}
\textbf{Feature Extractor $\phi$} & \textbf{LPIPS} $\downarrow$ & \textbf{DreamSim} $\downarrow$ & \textbf{CLIP-Sim} $\uparrow$ & \textbf{Quality} $\uparrow$
& \textbf{LPIPS} $\downarrow$ & \textbf{DreamSim} $\downarrow$ & \textbf{CLIP-Sim} $\uparrow$ & \textbf{Quality} $\uparrow$ \\
\midrule
VAE latent &
\textbf{0.240} & \textbf{0.066} & \textbf{0.955} & \textbf{0.828} &
0.305 & \underline{0.058} & \textbf{0.952} & \textbf{0.821}\\
DINOv2~\cite{oquab2024dinov2} &
0.336 & 0.122 & 0.926 & 0.812 &
\underline{0.296} & 0.068 & 0.943 & 0.798\\
CLIP~\cite{radford2021learning} &
\underline{0.297} & \underline{0.107} & \underline{0.931} & \underline{0.821} &
\textbf{0.288} & \textbf{0.057} & \underline{0.952} & \underline{0.810}\\
\bottomrule
\end{tabular}
}

\end{table*}

\begin{table}[t]
\centering
\caption{\textbf{Runtime overhead of different feature extractors.} Using DINOv2 or CLIP requires decoding each block back to RGB and running the vision backbone, adding significant per-block overhead. VAE latent features avoid this cost entirely by reusing the already-available encoder output. Measured on a single A100 GPU at 480p resolution.}
\label{tab:abl_phi_runtime}
\begin{tabular}{lccc}
\toprule
\textbf{Feature Extractor $\phi$} & \textbf{VAE Decode} & \textbf{Backbone} & \textbf{Total (20s video)} \\
\midrule
VAE latent & --- & --- & 0.3~ms \\
DINOv2~\cite{oquab2024dinov2} & 17.4~ms & 62.2~ms & 2.3~s \\
CLIP~\cite{radford2021learning} & 17.4~ms & 14.7~ms & 0.9~s \\
\bottomrule
\end{tabular}

\end{table}

\section{Feature Extractor Comparison}
\label{sec:supp_feature_extractor}
In Section 3.2 of the main paper, we noted that VAE latent features outperform alternative feature extractors for correspondence matching. Here we provide the full comparison.

Table~\ref{tab:abl_phi} compares three choices for the feature extractor $\phi$ used for computing the patch descriptor: VAE encoder's latent representation, pretrained DINOv2~\cite{oquab2024dinov2}, and CLIP~\cite{radford2021learning}.
VAE latent features achieve the best reconstruction on VACE-Bench and competitive performance on DL3DV, without requiring an external feature model.

Table~\ref{tab:abl_phi_runtime} reports the corresponding runtime overhead.
Using DINOv2 or CLIP requires decoding each block's latents back to RGB via the VAE decoder and then running the vision backbone, adding 79.7~ms and 32.2~ms per block respectively.
Over a 20-second video (29 blocks), this amounts to 2.3~seconds for DINOv2 and 0.9~seconds for CLIP.
VAE latent features incur no additional cost since they are already available from the encoding step.

\section{Dataset Details and Preprocessing}
\label{sec:supp_dataset}

\myparagraph{VACE-Benchmark (ABAB).}
We follow the standard VACE-Benchmark~\cite{vace} evaluation protocol for video preprocessing: each source video is uniformly sampled to 81 frames and resized to $480{\times}832$ with center cropping.
We retain all source videos with at least 81 frames.
To evaluate long-range consistency, we follow the concatenation protocol of~\cite{guo2026efficient} and construct 100 evaluation sequences (with a fixed random seed) by randomly pairing two source clips and concatenating them in an A$\rightarrow$B$\rightarrow$A$\rightarrow$B order, yielding $4{\times}81{=}324$ frames per sequence at 16\,fps (${\sim}$20\,seconds).
The repeated structure produces ground-truth correspondence pairs, where each pair maps a frame in the first occurrence of a clip to the same frame in its second occurrence.
Conditioning signals (depth maps, optical flow, and scribble maps) are extracted using the official VACE annotators applied to the concatenated sequence.
Text captions are obtained from the original VACE-Benchmark English prompts and stylized into five artistic variants per clip using Qwen2.5-14B-Instruct.

\myparagraph{DL3DV.}
We use the DL3DV-10K dataset~\cite{ling2024dl3dv} to evaluate long-range consistency under controlled, geometry-grounded viewpoint revisits.
We select scenes spanning diverse indoor and outdoor environments from the DL3DV-GS-960P split, which provides calibrated COLMAP camera poses and pre-trained 3D Gaussian Splatting (3DGS) models~\cite{kerbl3Dgaussians}.
For each scene, we construct a 60-second looping video (960 frames at 16\,fps) as follows.
We first extract the first loop from the camera trajectory by finding the point where the camera first returns closest to its starting position after traveling at least 25\% of the total path length.
The last 20 cameras are smoothly blended toward the initial camera (linear interpolation for translation, SLERP for rotation) to close the loop.
The overlap segment yields ground-truth correspondence pairs per scene, with each pair separated by approximately 48 seconds.
Conditioning signals are extracted using the same VACE annotator pipeline as VACE-Benchmark.
Text captions are generated by captioning the first rendered frame with Qwen2.5-VL-7B-Instruct and stylizing into five variants with Qwen2.5-14B-Instruct.
This dataset contains 180 samples in total.

\section{Per-Metric V-Bench Scores}
\label{sec:supp_vbench}

Tables in the main paper report the average V-Bench quality score. Table~\ref{tab:supp_vbench} provides the individual metric breakdown for all methods on both benchmarks.
Table~\ref{tab:supp_vbench_orthogonal} provides the corresponding per-metric breakdown for the distillation method sensitivity experiment.

\begin{table*}[t]
\centering
\caption{\textbf{Per-metric V-Bench quality scores.} We report individual normalized V-Bench~\cite{huang2023vbench} metrics for all methods on both benchmarks. The rightmost column shows the average across all six metrics (reported as the aggregate quality score in the main paper). Best among autoregressive methods is \textbf{bold}, second best is \underline{underlined}.}
\label{tab:supp_vbench}
\vspace{1mm}
\resizebox{\textwidth}{!}{%
\begin{tabular}{llccccccc}
\toprule

\multicolumn{9}{c}{\textbf{VACE-Bench}} \\
& & Subj.$\uparrow$ & BG$\uparrow$ & Flicker$\uparrow$ & Motion$\uparrow$ & Aesth.$\uparrow$ & Imaging$\uparrow$ & \textbf{Avg.}$\uparrow$ \\
\midrule
\multirow{4}{*}{\shortstack[l]{Offline}} & {\color{gray} VACE 1.3B} & {\color{gray} 0.929} & {\color{gray} 0.935} & {\color{gray} 0.934} & {\color{gray} 0.962} & {\color{gray} 0.606} & {\color{gray} 0.643} & {\color{gray} 0.835} \\
 & {\color{gray} Control-A-Video} & {\color{gray} 0.668} & {\color{gray} 0.799} & {\color{gray} 0.875} & {\color{gray} 0.878} & {\color{gray} 0.458} & {\color{gray} 0.657} & {\color{gray} 0.722} \\
 & {\color{gray} VideoComposer} & {\color{gray} 0.761} & {\color{gray} 0.864} & {\color{gray} 0.853} & {\color{gray} 0.852} & {\color{gray} 0.475} & {\color{gray} 0.537} & {\color{gray} 0.724} \\
 & {\color{gray} ControlVideo} & {\color{gray} 0.799} & {\color{gray} 0.867} & {\color{gray} 0.911} & {\color{gray} 0.927} & {\color{gray} 0.573} & {\color{gray} 0.591} & {\color{gray} 0.778} \\
\hdashline
\multirow{5}{*}{\shortstack[l]{Auto- \\ regressive}} & Self-Forcing (4 steps) & 0.779 & 0.835 & 0.914 & 0.944 & \textbf{0.623} & 0.640 & 0.789 \\
 & LongLive (4 steps) & 0.873 & 0.891 & 0.916 & 0.953 & \underline{0.619} & \underline{0.650} & \underline{0.817} \\
 & Rolling Forcing (5 steps) & 0.819 & 0.863 & \textbf{0.936} & \textbf{0.962} & 0.597 & 0.616 & 0.799 \\
\cdashline{2-9}
 & Ours (2 step) & \underline{0.883} & \underline{0.900} & 0.918 & 0.947 & 0.594 & 0.642 & 0.814 \\
 & Ours (5 step) & \textbf{0.888} & \textbf{0.903} & \underline{0.931} & \underline{0.958} & 0.601 & \textbf{0.687} & \textbf{0.828} \\
\midrule

\multicolumn{9}{c}{\textbf{DL3DV}} \\
& & Subj.$\uparrow$ & BG$\uparrow$ & Flicker$\uparrow$ & Motion$\uparrow$ & Aesth.$\uparrow$ & Imaging$\uparrow$ & \textbf{Avg.}$\uparrow$ \\
\midrule
\multirow{4}{*}{\shortstack[l]{Offline}} & {\color{gray} VACE 1.3B} & {\color{gray} 0.924} & {\color{gray} 0.928} & {\color{gray} 0.886} & {\color{gray} 0.948} & {\color{gray} 0.587} & {\color{gray} 0.644} & {\color{gray} 0.819} \\
 & {\color{gray} Control-A-Video} & {\color{gray} 0.771} & {\color{gray} 0.857} & {\color{gray} 0.844} & {\color{gray} 0.858} & {\color{gray} 0.413} & {\color{gray} 0.619} & {\color{gray} 0.727} \\
 & {\color{gray} VideoComposer} & {\color{gray} 0.771} & {\color{gray} 0.864} & {\color{gray} 0.836} & {\color{gray} 0.850} & {\color{gray} 0.475} & {\color{gray} 0.514} & {\color{gray} 0.718} \\
 & {\color{gray} ControlVideo} & {\color{gray} 0.876} & {\color{gray} 0.909} & {\color{gray} 0.899} & {\color{gray} 0.919} & {\color{gray} 0.583} & {\color{gray} 0.580} & {\color{gray} 0.794} \\
\hdashline
\multirow{5}{*}{\shortstack[l]{Auto- \\ regressive}} & Self-Forcing (4 steps) & 0.899 & 0.912 & 0.885 & 0.933 & 0.559 & 0.550 & 0.790 \\
 & LongLive (4 steps) & 0.914 & 0.914 & 0.878 & 0.945 & 0.567 & \underline{0.618} & \underline{0.806} \\
 & Rolling Forcing (5 steps) & 0.904 & 0.908 & \textbf{0.919} & \textbf{0.956} & \underline{0.580} & 0.568 & 0.806 \\
\cdashline{2-9}
 & Ours (2 step) & \textbf{0.917} & \textbf{0.915} & 0.875 & 0.936 & 0.565 & 0.584 & 0.799 \\
 & Ours (5 step) & \underline{0.915} & \underline{0.915} & \underline{0.899} & \underline{0.953} & \textbf{0.590} & \textbf{0.653} & \textbf{0.821} \\
\bottomrule
\end{tabular}
}
\end{table*}

\begin{table*}[t]
\centering
\caption{\textbf{Per-metric V-Bench quality scores for each distillation method.} We report individual normalized V-Bench~\cite{huang2023vbench} metrics for each autoregressive distillation backbone with and without our correspondence mechanism. Adding correspondences maintains comparable overall visual quality across different autoregressive backbones, with some variations in individual metrics.}
\label{tab:supp_vbench_orthogonal}
\vspace{1mm}
\resizebox{\textwidth}{!}{%
\begin{tabular}{lccccccc}
\toprule

\multicolumn{8}{c}{\textbf{VACE-Bench}} \\
& Subj.$\uparrow$ & BG$\uparrow$ & Flicker$\uparrow$ & Motion$\uparrow$ & Aesth.$\uparrow$ & Imaging$\uparrow$ & \textbf{Avg.}$\uparrow$ \\
\midrule
Self-Forcing & 0.779 & 0.835 & 0.914 & 0.944 & 0.623 & 0.640 & 0.789 \\
\quad + Correspondence & 0.849 & 0.877 & 0.912 & 0.941 & 0.601 & 0.581 & 0.794 \\
\hdashline
LongLive & 0.873 & 0.891 & 0.916 & 0.953 & 0.619 & 0.650 & 0.817 \\
\quad + Correspondence & 0.875 & 0.894 & 0.906 & 0.935 & 0.603 & 0.628 & 0.807 \\
\hdashline
Rolling Forcing & 0.819 & 0.863 & 0.936 & 0.962 & 0.597 & 0.616 & 0.799 \\
\quad + Correspondence & 0.888 & 0.903 & 0.931 & 0.958 & 0.601 & 0.687 & 0.828 \\
\midrule

\multicolumn{8}{c}{\textbf{DL3DV}} \\
& Subj.$\uparrow$ & BG$\uparrow$ & Flicker$\uparrow$ & Motion$\uparrow$ & Aesth.$\uparrow$ & Imaging$\uparrow$ & \textbf{Avg.}$\uparrow$ \\
\midrule
Self-Forcing & 0.899 & 0.912 & 0.885 & 0.933 & 0.559 & 0.550 & 0.790 \\
\quad + Correspondence & 0.914 & 0.910 & 0.873 & 0.938 & 0.533 & 0.483 & 0.775 \\
\hdashline
LongLive & 0.914 & 0.914 & 0.878 & 0.945 & 0.567 & 0.618 & 0.806 \\
\quad + Correspondence & 0.912 & 0.915 & 0.872 & 0.928 & 0.548 & 0.539 & 0.786 \\
\hdashline
Rolling Forcing & 0.904 & 0.908 & 0.919 & 0.956 & 0.580 & 0.568 & 0.806 \\
\quad + Correspondence & 0.915 & 0.915 & 0.899 & 0.953 & 0.590 & 0.653 & 0.821 \\
\bottomrule
\end{tabular}
}
\end{table*}

\clearpage
\bibliographystyle{splncs04}
\bibliography{main}

\end{document}